\documentclass{article} 
\usepackage{iclr2024_conference,times}

\usepackage{amsmath,amsfonts,bm}

\def\eqref#1{equation~\ref{#1}}

\def\1{\bm{1}}

\DeclareMathAlphabet{\mathsfit}{\encodingdefault}{\sfdefault}{m}{sl}
\SetMathAlphabet{\mathsfit}{bold}{\encodingdefault}{\sfdefault}{bx}{n}

\usepackage{hyperref}
\usepackage{url}
\usepackage{graphicx}
\usepackage{amsmath,amsthm,amssymb}
\usepackage{mathrsfs}
\usepackage{subcaption}
\usepackage{cleveref}
\usepackage{wrapfig}
\usepackage{multirow}
\usepackage{colortbl}
\usepackage{booktabs}
\usepackage{bbding}
\usepackage[normalem]{ulem}
\useunder{\uline}{\ul}{}
\definecolor{colorcite}{RGB}{15,90,110}
\definecolor{newcontent}{RGB}{232,79,10}
\hypersetup{
hidelinks,
colorlinks=true,
linkcolor=red,
citecolor=colorcite,
urlcolor = black
}

\theoremstyle{definition}
\newtheorem{definition}{Definition}
\newtheorem{proposition}{Proposition}

\title{EventRPG: Event Data Augmentation with Relevance Propagation Guidance}

\author{Mingyuan Sun\textsuperscript{\rm 1,\,2}, Donghao Zhang\textsuperscript{\rm 3}, Zongyuan Ge\textsuperscript{\rm 4}, Jiaxu Wang\textsuperscript{\rm 1}, Jia Li\textsuperscript{\rm 5},\\
\textbf{Zheng Fang\textsuperscript{\rm 2,}\footnotemark[1]\ \ \& Renjing Xu\textsuperscript{\rm 1,}}\thanks{Corresponding authors.} \\
\textsuperscript{\rm 1}The Hong Kong University of Science and Technology (Guangzhou)\quad
\textsuperscript{\rm 2}Northeastern University\\
\textsuperscript{\rm 3}Seeing Machines \quad
\textsuperscript{\rm 4}Monash University \quad
\textsuperscript{\rm 5}Peking University\\
\small\texttt{mingyuansun20@gmail.com,\,peter.zhang1@seeingmachines.com,} \\
\small\texttt{zongyuan.ge@monash.edu,\,jwang457@connect.hkust-gz.edu.cn,} \\
\small\texttt{j.gaga.lee@gmail.com,\,fangzheng@mail.neu.edu.cn,\,renjingxu@hkust-gz.edu.cn}
}

\iclrfinalcopy 
\begin{document}

\maketitle

\begin{abstract}
Event camera, a novel bio-inspired vision sensor, has drawn a lot of attention for its low latency, low power consumption, and high dynamic range. Currently, overfitting remains a critical problem in event-based classification tasks for Spiking Neural Network (SNN) due to its relatively weak spatial representation capability. Data augmentation is a simple but efficient method to alleviate overfitting and improve the generalization ability of neural networks, and saliency-based augmentation methods are proven to be effective in the image processing field. However, there is no approach available for extracting saliency maps from SNNs. Therefore, for the first time, we present Spiking Layer-Time-wise Relevance Propagation rule (\texttt{SLTRP}) and Spiking Layer-wise Relevance Propagation rule (\texttt{SLRP}) in order for SNN to generate stable and accurate CAMs and saliency maps. Based on this, we propose \texttt{EventRPG}, which leverages relevance propagation on the spiking neural network for more efficient augmentation. Our proposed method has been evaluated on several SNN structures, achieving state-of-the-art performance in object recognition tasks including N-Caltech101, CIFAR10-DVS, with accuracies of $85.62\%$ and $85.55\%$, as well as action recognition task SL-Animals with an accuracy of $91.59\%$. Our code is available at \href{https://github.com/myuansun/EventRPG}{\textcolor[HTML]{ED008A}{\texttt{https://github.com/myuansun/EventRPG}}}.
\end{abstract}
\vspace{-0.4cm}
\section{Introduction}
\vspace{-0.3cm}
With the advent of event cameras, researchers have focused on applying the brain-inspired technique to achieve a variety of tasks, as the asynchronous nature of event cameras mimics the way the biological visual system works~\citep{gallego2020eventsurvey}. Event cameras record the change in brightness of each pixel, and once the change in brightness of a pixel exceeds a predetermined threshold, an event is triggered. The intrinsic properties of event cameras give them several advantages over RGB cameras, including low power consumption, high dynamic range, low latency, and high temporal resolution. These benefits highlight the potential of event cameras in challenging scenarios, such as low-light and high-speed conditions, which has led to some research emphasizing the use of event cameras for robotic sensing in challenging situations~\citep{devo, chen2023esvio}. Spiking Neural Network (SNN)~\citep{maass1997networks} is a type of neural network that is inspired by the way biological neurons communicate with each other. By integrating the biological neuronal dynamics into individual neurons, SNN becomes capable of representing intricate spatio-temporal information and dealing with asynchronous data naturally, typically event-based data.

In terms of classification tasks, a number of event-based datasets, such as N-MNIST, N-Caltech101~\citep{orchard2015converting}, and CIFAR10-DVS~\citep{li2017cifar10}, have been used to evaluate the performance of artificial neural networks (ANNs) and SNNs. However, the issue of overfitting still poses a significant challenge for event-based datasets. Data augmentation is an efficient method for improving the generalization and performance of a model. Lots of methods have been proposed to augment event-based data, for example, transferring classic geometric augmentations from image field to event-based field~\citep{NDA}, randomly dropping events~\citep{eventdrop}, and mixing two event streams with a randomly sampled mask~\citep{shen2023eventmix}. Nevertheless, current mixing augmentation strategies in event-based field do not consider the size and location information of label-related objects, and thus may produce events with incorrect labels and disrupt the training process. To address this problem in image processing field, \citet{uddin2020saliencymix, kim2020puzzle} mix the label-related objects together based on the saliency information obtained from neural networks. This paradigm achieves better results compared with conventional non-saliency augmentations. \citet{kim2021visual} was the first to explore acquiring saliency information from SNNs. In this work, Spiking Activation Map (SAM) was presented to reveal the model's attention by weighted adding of the intermediate feature maps. Since the CAMs obtained from this method are not related to the network's predictions, their precision regarding the shape and position of label-related objects is still not satisfactory. 

In this paper, we present two novel methods extended from layer-wise relevance propagation (LRP)~\citep{bach2015pixel} to visualize the label-related saliency information of SNNs, each offering distinct advantages in terms of temporal precision and computational time. Moreover, guided by this saliency information, we develop two data augmentation approaches targeted at event-based data, demonstrating significant improvements in the performance and generalization capability of SNNs across multiple classification tasks.

Our contributions are summarized as follows:
\begin{itemize}
\item We propose Spiking Layer-Time-wise Relevance Propagation (\texttt{SLTRP}) and Spiking Layer-wise Relevance Propagation (\texttt{SLRP}) to accurately reveal the saliency information of SNNs. The former reveals information across time, while the latter is less time-consuming.
\item We present RPGDrop and RPGMix. By dropping and mixing events with the guidance of relevance propagation obtained from \texttt{SLRP} or \texttt{STLRP}, the augmented samples exhibit increased diversity and tight correlation with the labels. Combined with several geometric data augmentations, we formulate our data augmentation strategy namely \texttt{EventRPG}.
\item We evaluate our proposed saliency visualizing method and data augmentation method using various SNNs on event-based object and action recognition datasets. Experiments demonstrate that our \texttt{SLRP} and \texttt{SLTRP} can generate high quality CAMs and saliency maps with sub-optimal computing time. \texttt{EventRPG} achieves state-of-the-art performance on both object recognition and action recognition tasks with limited time consumption.
\end{itemize}

\vspace{-0.5cm}
\section{Preliminary}
\vspace{-0.3cm}
Layer-wise Relevance propagation (LRP) was first introduced in \citep{bach2015pixel} as a visualization tool for generating saliency maps that show the contribution of individual pixels in the input data to the model prediction or a specific class, facilitating the interpretability of neural networks. According to the LRP rule, we assign a value to each neuron to represent the neuron's contribution to the prediction or target class, called Relevance Score. The idea of LRP is to find a propagation rule satisfying the following definition.

\begin{definition}[Conservation Property]
\label{def:1}
On a neural network with $L$ layers, the relevance score of each layer satisfies
\begin{equation}
    c = \sum_{i}R^{(l)}_{i}, \qquad \forall l \in [0,L],
    \label{eq:1}
\end{equation}
where $c$ is a constant value and $R^{(l)}_{i}$ is the relevance score of $i^{th}$ neuron on layer $l$. $R^{(0)}$ is the relevance score before input layer.
\end{definition}
\vspace{-0.3cm}
We calculate the relevance score of $i^{th}$ input neuron on layer $l$ $R^{(l-1)}_{i}$ by the sum of all relevance scores propagated from all the connections in this layer:
\begin{equation}
R_{i}^{(l-1)}=\sum_{j}R_{i \leftarrow j}^{(l-1, l)}.
    \label{eq:2}
\end{equation}
Utilizing the $\alpha\beta$-rule~\citep{montavon2017explaining} and abandoning the bias term $b$, we obtain the relevance score that should be propagated from the $j^{th}$ neuron in layer $l$ to the $i^{th}$ neuron in layer $l-1$:
\begin{equation}
R_{i \leftarrow j}^{(l-1, l)}=R_{j}^{(l)} \cdot\biggl(\alpha \cdot \frac{z_{i j}^{+}}{\sum_{i}z_{i j}^{+}}+\beta \cdot \frac{z_{i j}^{-}}{\sum_{i}z_{i j}^{-}}\biggr),
\label{eq:3}
\end{equation}
where $\alpha + \beta = 1$.
$z^+_{ij} = w_{i j}^{+} \cdot x_{i}^{+} + w_{i j}^{-} \cdot x_{i}^{-}$ and $z^-_{ij} = w_{i j}^{+} \cdot x_{i}^{-} + w_{i j}^{-} \cdot x_{i}^{+}$ are respectively the positive and negative contribution of the $i^{th}$ neuron in layer $l-1$ to the $j^{th}$ neuron in layer $l$. This relevance propagation rule satisfies \cref{def:1}, and the proof can be seen in \cref{sec:ab_proof}.

\vspace{-0.3cm}
\section{Spiking Relevance Propagation Rule}
\vspace{-0.3cm}
In an SNN, information is represented by the timing and frequency of spikes. Neurons in an SNN generate spikes when their membrane voltage reaches a certain threshold. These spikes propagate through the network, influencing the activity of other neurons layer by layer. Two popular and fundamental models used to represent basic spiking neurons are the Leaky Integrate-and-Fire (LIF) neuron model and the Integrate-and-Fire (IF) neuron model:
\vspace{-0.2cm}
\begin{align}
\label{eq:4} {f}_{LIF}(V, I)&=e^{-\frac{\Delta t}{\tau}} V[t-1]+\left(1-e^{-\frac{\Delta t}{\tau}}\right) I[t],\\
\label{eq:5} {f}_{IF}(V, I)&=V[t-1]+I[t],
\end{align}
where $\tau$ is the attenuation factor and $\Delta t$ is the interval between time steps satisfying $\Delta t < < \tau$. Membrane voltage $V$ can be interpreted as the information reserved by a neuron from the previous time step $t-1$. $I[t]$ indicates the input of the neuron at the current time step.

\begin{figure}[t]
\centering
\includegraphics[width=\textwidth]{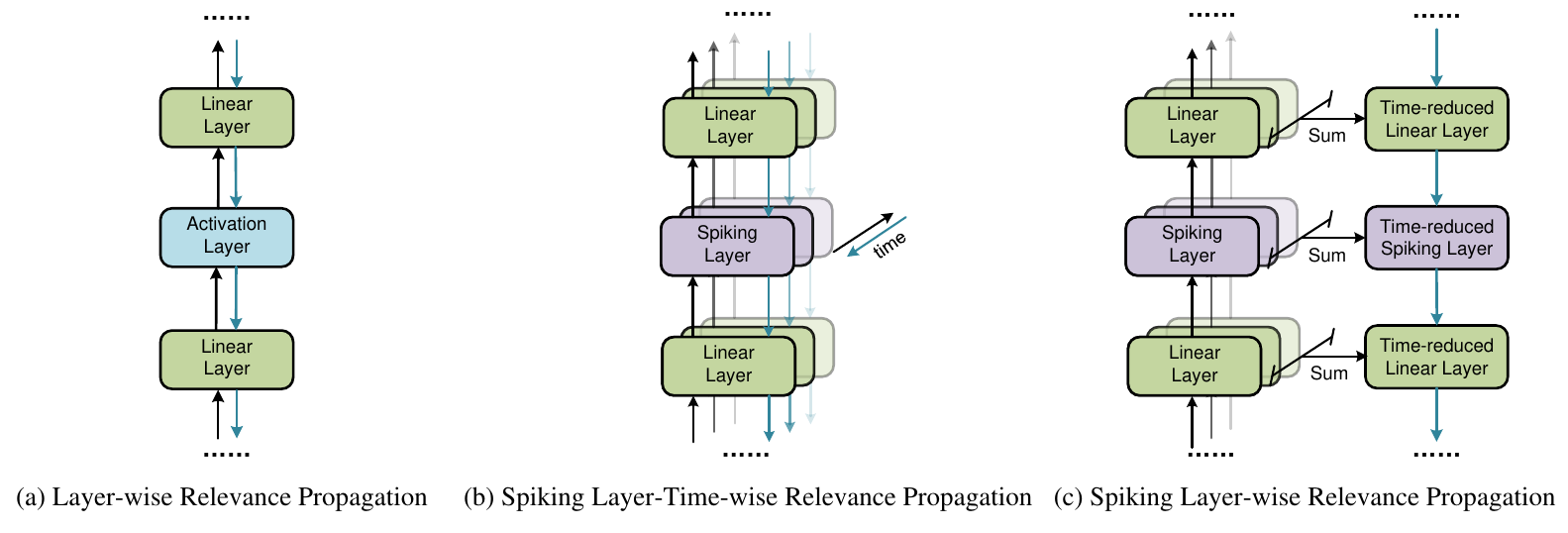}
\caption{Forward propagation flow and relevance propagation flow of ANNs (a) and SNNs (b, c).}
\label{fig:2}
\vspace{-0.5cm}
\end{figure}

In practice, researchers would replace the activation layer in an ANN with spiking layers represented by \cref{eq:4} or \cref{eq:5} to construct an SNN. As shown in \cref{fig:2}a, an ANN consists of linear layers and activation layers and passes information layer by layer. These two layer types do not include variables related to time and as such we propagate the relevance scores through them layer by layer. 
\vspace{-0.6cm}
\subsection{Relevance Propagation for Linear Layers in SNNs}
\vspace{-0.2cm}
\label{rp_linear}
Information in SNNs is propagated through time in spiking layers, while the linear layers do not transmit information across different time steps. Any linear layer's output $y[t]$ at time step $t$ only correlates to its input $x[t]$ at time step $t$, and we conduct relevance propagation using \cref{eq:3} on this layer separately for each time step. Specifically, for linear layer $l$, we extend the \cref{eq:3} with suffix ``$[t]$" representing the time step $t$, and we have
\begin{align}
R_{i}^{(l-1)}[t]&=\sum_{j}R_{i \leftarrow j}^{(l-1, l)}[t],
\label{eq:6}\\
R_{i \leftarrow j}^{(l-1, l)}[t]&=R_{j}^{(l)}[t] \cdot\biggl(\alpha \cdot \frac{z_{i j}^{+}[t]}{\sum_{i}z_{i j}^{+}[t]}+\beta \cdot \frac{z_{i j}^{-}[t]}{\sum_{i}z_{i j}^{-}[t]}\biggr),
\label{eq:7}
\end{align}
in which $z^+_{ij}[t] = w_{i j}^{+} \cdot x_{i}^{+}[t] + w_{i j}^{-} \cdot x_{i}^{-}[t]$ and $z^-_{ij}[t] = w_{i j}^{+} \cdot x_{i}^{-}[t] + w_{i j}^{-} \cdot x_{i}^{+}[t]$.
Next, we focus on the derivation of the relevance propagation rule on spiking layers.

\vspace{-0.2cm}
\subsection{Spiking Layer-Time-wise Relevance Propagation}
\vspace{-0.1cm}
From \cref{eq:4} we know that in the forward propagation, the output of a LIF neuron at time step t ($t > 0$) depends on this layer's input current $I[t]$ at the current time step and the membrane voltage $V[t-1]$ from the previous time step. Therefore, we should propagate the relevance score to the neuron's internal voltage at the previous time step and to the input current at the current time step, as shown in \cref{fig:2}b. Assume there are overall T time steps. Similar to $\alpha\beta$-rule, we propagate the relevance score based on its positive contribution and negative contribution at each time step $t\in[1,T]$. Consider a neuron in spiking layer $l$ at time step t. We decomposite and represent it in a general manner:
\begin{equation}
f(V, I)=c \cdot V[t-1]+d \cdot I[t]
\label{eq:8}
\end{equation}
\vspace{-0.2cm}
where c and d are coefficients depending on the neuron type:
\begin{align*}
\begin{aligned}
c = \begin{cases}e^{-\frac{\Delta t}{\tau}} & \text{LIF Neuron}, \\ 1 & \text{IF Neuron},\end{cases}
\end{aligned}
\qquad
\begin{aligned}
d = \begin{cases}1-e^{-\frac{\Delta t}{\tau}} & \text{LIF Neuron}, \\ 1 & \text{IF Neuron}.\end{cases}
\end{aligned}
\end{align*}

\vspace{-0.2cm}
Based on \cref{eq:8}, at time step t, we define the proportion of relevance score that should be propagated to the previous time step as
\vspace{-0.2cm}
\begin{equation}
\gamma[t] = \alpha \cdot \frac{c\cdot V^+[t-1]}{c\cdot V^+[t-1] + d \cdot I^+[t]} + \beta \cdot \frac{c\cdot V^-[t-1]}{c\cdot V^-[t-1] + d \cdot I^-[t]},
\label{eq:9}
\end{equation}
where superscripts $^+$ and $^-$ denote the positive and negative values, respectively.
For each neuron in spiking layer $l$, By initializing the relevance score at the final time step as $R^{(l-1)}[T] = R^{(l)}[T]$ and subsequently updating relevance scores iteratively as time step $t$ from $T$ to $1$ using
\begin{align}
\label{eq:10} R^{(l-1)}[t-1]&\leftarrow\gamma[t] \cdot R^{(l-1)}[t] + R^{(l)}[t-1],\\
\label{eq:11} R^{(l-1)}[t]&\leftarrow(1-\gamma[t])\cdot R^{(l-1)}[t],
\end{align}
we propagate the relevance score of every neuron in spiking layers to all time steps. With iteration formulas (\ref{eq:10}) and (\ref{eq:11}), we could derive the relevance score at each time step
\begin{equation}
\label{eq:12}
R^{(l-1)}[t]=(1-\gamma[t])\left(\sum^T_{i=t+1}R^{(l)}[i]\prod_{j=t+1}^{i}\gamma[j]+R^{(l)}[t]\right).
\end{equation}

\begin{proposition}
\label{prop:1}
The sum of relevance scores propagated after spiking layer $l$ from time step $1$ to $k (k<T)$ is
\vspace{-0.2cm}
\begin{equation}
\label{eq:13}
\sum^{k}_{t=1} R^{(l-1)}[t]=\sum^{k}_{t=1} R^{(l)}[t] + \sum^T_{i=k+1}R^{(l)}[i]\prod_{j=k+1}^{i}\gamma[j].
\end{equation}
\end{proposition}
The proof is provided in \cref{sec:prop1proof}. With \cref{prop:1}, we have
\vspace{-0.1cm}
\begin{align}
\notag \sum^{T}_{t=1} R^{(l-1)}[t]=\sum^{T-1}_{t=1} R^{(l-1)}[t] + R^{(l-1)}[T] &=\sum^{T-1}_{t=1} R^{(l)}[t] + \gamma[T]R^{(l)}[T] + (1-\gamma[T])R^{(l)}[T]\\
&=\sum^{T-1}_{t=1} R^{(l)}[t] + R^{(l)}[T]=\sum^{T}_{t=1} R^{(l)}[t].
\label{eq:14}
\end{align}
This is a stronger Conservation Property since the relevance score stays unchanged for every neuron in a spiking layer. This indicates that leveraging formulas (\ref{eq:10}) and (\ref{eq:11}), we are able to propagate relevance scores through any spiking layer while satisfying the Conservation Property. Combining with relevance propagation rules introduced in \cref{rp_linear}, we could propagate relevance scores to any layer in an SNN, thereby revealing the saliency information across time, namely Spiking Layer-Time-wise Relevance Propagation~(\texttt{SLTRP}).
\vspace{-0.3cm}
\subsection{Spiking Layer-wise Relevance Propagation}
\vspace{-0.2cm}
\texttt{SLTRP} could reveal the saliency information at any time step, and would thus cost more time to conduct the whole relevance propagation process compared with ANNs. Under some circumstances, e.g., on datasets transformed from static images where coordinates of events tend to be fixed w.r.t. the time, we don't need to obtain the saliency score of some specific time step and only require the saliency information stacked from all time steps $R^{(l)}\equiv\frac{1}{T}\sum_t^T R^{(l)}[t]$. For any spiking layer $l$, from \cref{eq:14} we have $R^{(l-1)}=R^{(l)}$. 

In terms of linear layers, we sum the positive and negative contribution values of the $i^{th}$ neuron to $j^{th}$ neuron through time dimension as
\begin{align*}
z^+_{ij}&\equiv\frac{1}{T}\sum_{t}z^+_{ij}[t] = \frac{1}{T}\sum_{t}(w_{i j}^{+} \cdot x_{i}^{+}[t] + w_{i j}^{-} \cdot x_{i}^{-}[t])=\frac{1}{T}(w_{i j}^{+} \cdot \sum_{t}x_{i}^{+}[t] + w_{i j}^{-} \cdot \sum_{t}x_{i}^{-}[t]),\\
z^-_{ij}&\equiv\frac{1}{T}\sum_{t}z^-_{ij}[t] = \frac{1}{T}\sum_{t}(w_{i j}^{+} \cdot x_{i}^{-}[t] + w_{i j}^{-} \cdot x_{i}^{+}[t])=\frac{1}{T}(w_{i j}^{+} \cdot \sum_{t}x_{i}^{-}[t] + w_{i j}^{-} \cdot \sum_{t}x_{i}^{+}[t]).
\end{align*}
\vspace{-0.4cm}

Then we propagate the relevance scores of linear layer using \cref{eq:2} and \cref{eq:3} with different definitions of $R^{(l-1)}$, $R^{(l)}$, $z^+_{ij}$ and $z^-_{ij}$. This enables a spiking relevance propagation process without consideration of the time dimension, saving time costs and being more practical for datasets transformed from static images, namely Spiking Layer-wise Relevance Propagation~(\texttt{SLRP}) (see \cref{fig:2}c).
\vspace{-0.4cm}
\section{Relevance Propagation Guided Event Data Augmentation}
\vspace{-0.3cm}
\begin{figure}[t!]
  \centering  
    \begin{subfigure}{0.17\textwidth}
      \centering   
      \includegraphics[width=\linewidth]{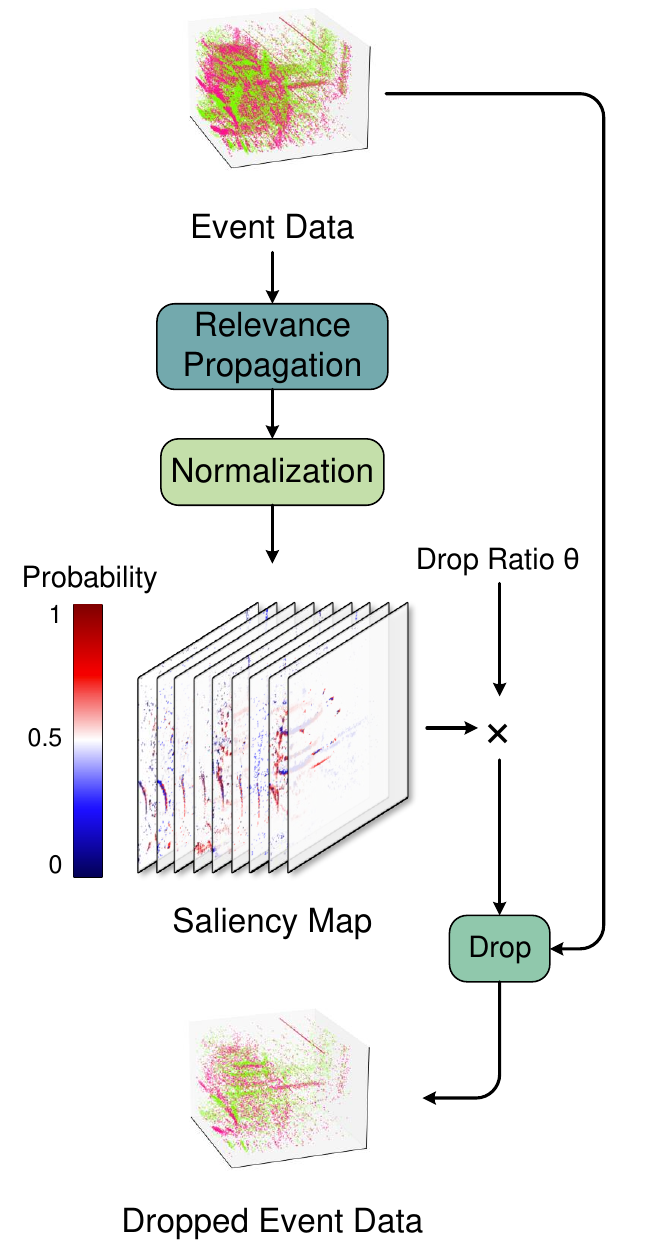}
        \caption{}
        \label{fig:rpgdrop}
    \end{subfigure}
    \begin{subfigure}{0.82\textwidth}
      \centering   
      \includegraphics[width=1\linewidth]{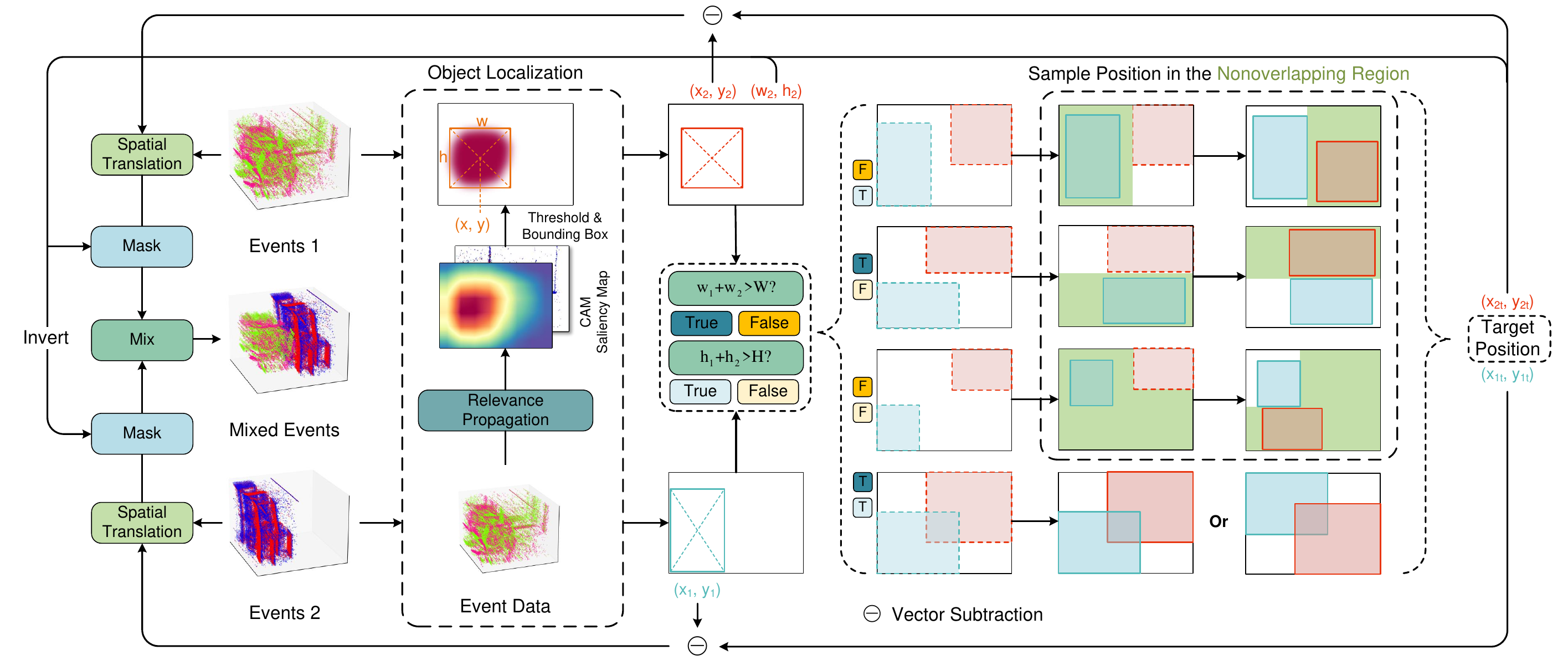}
        \caption{}
        \label{fig:rpgmix}
    \end{subfigure}
\vspace{-0.3cm}
\caption{
\label{fig:rpgaug}
Illustration of Relevance Propagation Guided Event Mix and Event Drop. \textbf{(a) RPGDrop}. Where the saliency map offers a higher value, events are more likely to be dropped.\textbf{(b) RPGMix}. }
\vspace{-0.7cm}
\end{figure}

\subsection{Saliency Map and Class Activation Map}
\vspace{-0.2cm}
In an SNN, we first leverage Contrastive Layer-wise Relevance Propagation (CLRP)~\citep{gu2018understanding} to initialize the relevance scores of the output layer for each time step. Then we propagate the relevance scores backward using \texttt{SLRP} or \texttt{SLTRP} depending on the dataset. Relevance scores are propagated throughout all layers to create saliency maps, while class activation maps (CAMs) can be formed in two ways. One method involves summing the relevance scores from a specific intermediate layer across the channel dimension, resulting in SLRP-CAM and SLTRP-CAM. Alternatively, CAMs can be generated by calculating a weighted sum between relevance scores and feature maps, a method referred to as SLRP-RelCAM and SLTRP-RelCAM~\citep{lee2021relevance}.
\vspace{-0.2cm}
\subsection{Relevance Propagation Guided Event Drop}
\vspace{-0.2cm}
\citet{eventdrop} has proven randomly dropping events to be an effective augmentation strategy. Furthermore, we expect to drop events more frequently in regions with label-related objects, motivated by the fact that disturbing regions with no label-related information (namely background) would have a negligible impact on the classifier's prediction. The label-related information can be provided by CAM and saliency map, where higher values imply higher model attention and label relevance. Since saliency map accurately reveals the relevance score of each pixel to the target in the input data, we leverage saliency map to guide dropping, detailly illustrated in \cref{fig:rpgdrop}. The higher the value of a pixel, the higher the probability that we will drop events on that pixel. $\theta$ is the parameter controlling the magnitude of augmentation.

\vspace{-0.2cm}
\subsection{Relevance Propagation Guided Event Mix}
\vspace{-0.2cm}
Event-based data, in contrast to image-based data, does not include color details, with the most crucial aspect being the texture information it contains. The overlapping of label-related objects will impair the texture details of these objects, which in turn further degrades the quality of features extracted in SNNs. Building upon this motivation, we propose Relevance Propagation Guided Event Mix (RPGMix). The whole mixing strategy is illustrated in \cref{fig:rpgmix}. For two event-based data candidates, we utilize relevance propagation to localize the label-related regions and obtain two bounding boxes. To mix two objects with clear texture features, we randomly select two positions ensuring minimal overlap of their bounding boxes. This involves initially positioning one box at a corner to maximize the nonoverlapping area for the other box's placement, then selecting positions for both boxes in order, maintaining minimal overlap and maximizing sampling options. Finally, the two event streams are moved to the sampled positions. Although this linear translation part prevents the overlapping of label-related objects, the background of one object would still overlap with the other object. Moreover, in one single time step, the representation ability of the spiking neurons (which only output binary information) is much worse than that of the activation layer (usually ReLU) of ANNs, making them less capable of spatial resolution and more likely to fall into local optima. Therefore, to promise the presence of only events from a single event stream candidate per pixel, avoiding regions with mixed information from interfering with the SNN, we adopt a CutMix strategy to mask the two event streams based on the bounding box of the second event stream, as demonstrated in the left part in \cref{fig:rpgmix}. \citet{kim2020puzzle} takes the sum of each sample's mask as the ratio of their corresponding labels. This ensures that the proportion of labels in the mixed label matches the proportion of pixels belonging to each sample. In our approach, we further aim to align the proportion of labels with the proportion of label-related pixels, which can be estimated using the bounding boxes. As a result, the labels of the two event streams are mixed as
\vspace{-0.1cm}
\begin{equation}
L_{mix} = \frac{L_{1}(w_{1}h_{1} - S_{overlap}) + L_{2}w_{2}h_{2}}{w_{1}h_{1} + w_{2}h_{2} - S_{overlap}},
\label{eq:15}
\end{equation}
\vspace{-0.5cm}

\begin{wrapfigure}[14]{t}{0.5\linewidth}
\centering
\includegraphics[width=\linewidth]{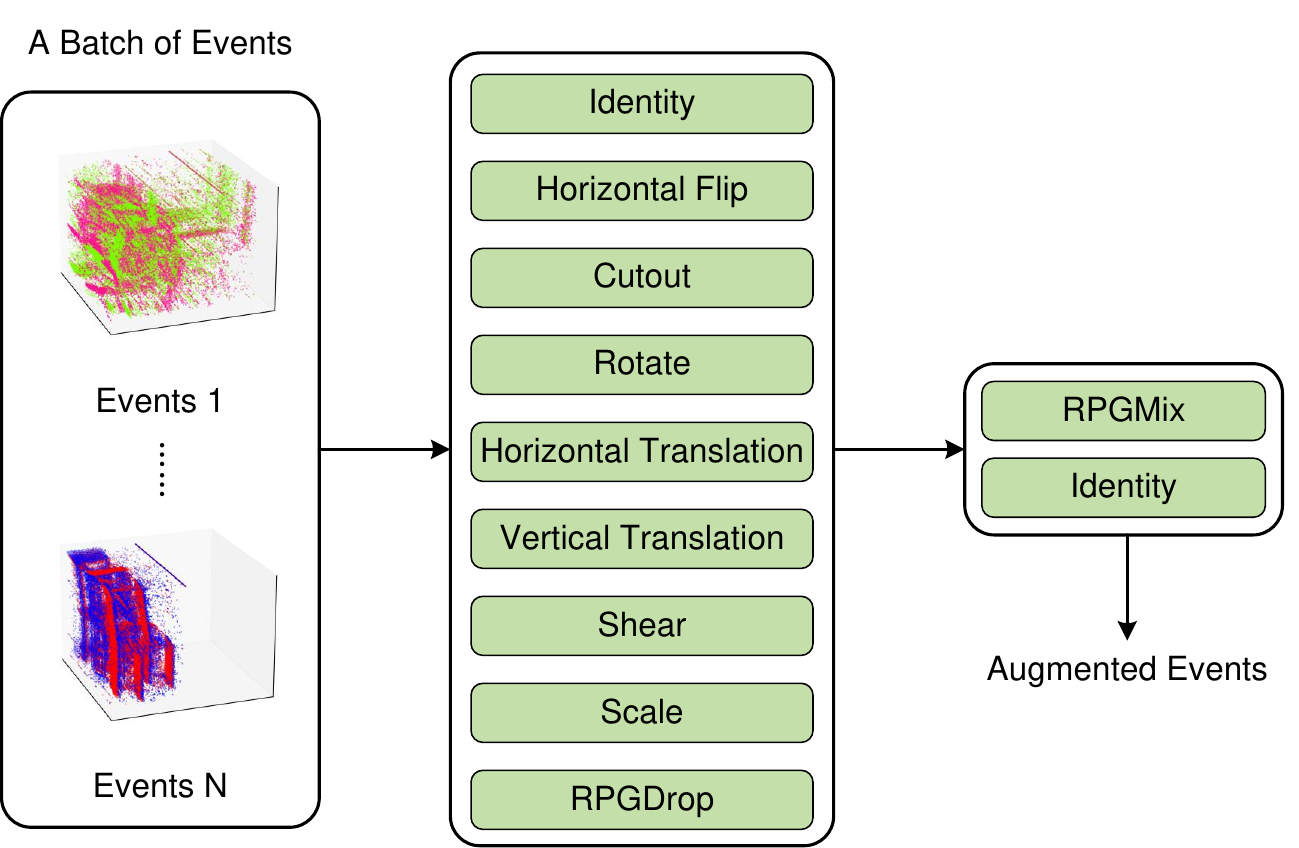}
\caption{Augmentation process of \texttt{EventRPG}.}
\label{fig:eventrpg}
\end{wrapfigure}
where $w_{i}$ and $h_{i}$ denote the width and height of the bounding box in the event stream $i$. $L_1$ and $L_2$ are the one-hot labels of the two event streams and $S_{overlap}$ is the area of the overlapping region of the two bounding boxes.

\vspace{-0.2cm}
\subsection{EventRPG}
\vspace{-0.2cm}
Combined with a few geometric data augmentation methods in NDA~\citep{NDA}, we formulate our data augmentation strategy namely \texttt{EventRPG} as shown in \cref{fig:eventrpg}. Specifically, for a batch of input event streams, each of the event streams would be augmented with randomly sampled policy and magnitude. They then have a probability of $0.5$ to be augmented by RPGMix.

\vspace{-0.4cm}
\section{Experiments}
\vspace{-0.3cm}
\subsection{Effectiveness of SLRP and SLTRP}
\vspace{-0.2cm}
In this subsection, we evaluate the effectiveness of our approach for generating CAMs and saliency maps. Current event-based datasets for classification can be divided into two tasks: object recognition task and action recognition task. Generally, the former mainly involves event streams generated from jittering of static images, while the latter mainly involves event streams recorded in real environments, containing more dynamic information. We perform experiments on both types of datasets to showcase the effectiveness of our approach. We visualize the feature before the last fully connected layer as CAM.

\vspace{-0.2cm}
\begin{table}[htbp]
  \centering
  \resizebox{\textwidth}{!}{
    \begin{tabular}{ccccccccccc} 
      \toprule
      \multirow{3}{*}{Method}                 & \multicolumn{6}{c}{Object Recognition}                                                                                      & \multicolumn{4}{c}{Action Recognition}                                             \\ 
      \cmidrule(r){2-7} \cmidrule(r){8-11}
                                              & \multicolumn{2}{c}{N-Caltech101}        & \multicolumn{2}{c}{CIFAR10-DVS}         & \multicolumn{2}{c}{N-Cars}              & \multicolumn{2}{c}{DVSGesture}          & \multicolumn{2}{c}{SL-Animals}           \\ 
      \cmidrule(r){2-3} \cmidrule(r){4-5} \cmidrule(r){6-7} \cmidrule(r){8-9} \cmidrule(r){10-11}
                                              & A.I.~$\uparrow$ & A.D.~$\downarrow$ & A.I.~$\uparrow$ & A.D.~$\downarrow$ & A.I.~$\uparrow$ & A.D.~$\downarrow$ & A.I.~$\uparrow$ & A.D.~$\downarrow$ & A.I.~$\uparrow$ & A.D.~$\downarrow$  \\ 
      \midrule
      SAM \citep{kim2021visual}               & 0.86              & 22.33               & 1.23              & 46.83               & 5.24              & \uline{5.89}        & 8.27              & 10.83               & 20.60             & \textbf{8.23}        \\
      Grad-CAM \citep{selvaraju2017grad}      & 12.03             & 41.71               & 8.31              & 22.15               & 21.77             & 17.74               & 0.41              & 67.09               & 0.80              & 81.44                \\
      Grad-CAM++ \citep{chattopadhay2018grad} & 24.82             & 10.32               & 6.23              & 26.33               & \textbf{26.04}    & \textbf{5.41}       & 7.99              & 10.41       & 11.78             & 22.21                \\ 
      \midrule
      SLRP-RelCAM \citep{lee2021relevance}         & 17.51             & 10.44               & 7.60              & 23.41               & 22.05             & 19.00               & 13.11             & \uline{6.41}       & \textbf{26.49}    & \uline{10.67}        \\
      SLTRP-RelCAM \citep{lee2021relevance}   & 17.54 & 10.44                    & 7.60 & 23.41                    & 22.05 & 19.00              & 13.05 & \textbf{6.40}                   & \textbf{26.49} & \uline{10.67}                   \\
      SLRP-CAM                                & \textbf{34.24}    & 5.75                & 8.41              & 23.50               & 15.72             & 23.80               & 7.79              & 29.99               & 10.88             & 40.95                \\
      SLTRP-CAM                               & \textbf{34.24}    & 5.75                & 8.41              & 23.50               & 15.72             & 23.80               & 7.79              & 30.02               & 10.88             & 40.95                \\
      SLRP-Saliency Map                       & 34.12             & \textbf{4.18}       & \uline{9.44}      & \uline{19.99}       & \uline{22.98}     & 6.77                & \textbf{22.20}    & 13.86               & \uline{22.30}     & 20.57                \\
      SLTRP-Saliency Map                      & \uline{34.17}     & \uline{4.19}        & \textbf{9.51}     & \textbf{19.98}      & 22.84             & 6.75                & \uline{21.17}     & 13.95               & \uline{22.30}     & 20.79                \\
      \bottomrule
      \end{tabular}
  }
  \caption{Comparison of A.I. and A.D. on event-based object recognition and action recognition datasets. We highlight the best results in bold and the second best results with underlining.}
  \label{tab:1}
  \vspace{-0.4cm}
  \end{table}
\vspace{-0.1cm}
\subsubsection{Objective Faithfulness}
\vspace{-0.2cm}
We adopt two widely used metrics, Average Drop (A.D.) and Average Increase (A.I.), to measure the objective faithfulness of our method compared with other typical visualization tools. These two metrics explain how well an attention map explains a model's attention by measuring the average change in the model's output when the attention map is applied as a mask to the input. The higher the A.I. and the lower the A.D., the better the attention map explains the model's attention. We compare our method with Grad-CAM~\citep{selvaraju2017grad}, Grad-CAM++~\citep{chattopadhay2018grad}, and SAM~\citep{kim2021visual} on both object recognition datasets and action recognition datasets.

On object recognition task, our method achieves the best performance in terms of A.I. and A.D. on large datasets including N-Caltech101 and CIFAR10DVS, demonstrating the effectiveness of our method in generating CAMs and saliency maps. On the N-Cars dataset, Grad-CAM++ outperforms our method. This disparity may be due to N-Cars being a binary classification task that solely focuses on detecting the presence of a car in the event stream, in contrast to other datasets requiring multi-class object localization.

In terms of action recognition task, methods with spiking relevance propagation outperform other methods significantly, with an A.I. metric almost thrice as good as the best method without spiking relevance propagation --- Grad-CAM++, in the DVSGesture dataset. In the SL-Animals dataset, RelCAM achieves the best results in terms of A.I. and also has a low A.D. value. Note that RelCAM is also obtained based on the relevance scores from \texttt{SLRP} and \texttt{SLTRP}. Therefore, its good results also helps to demonstrate the effectiveness of our spiking relevance propagation rules.
\vspace{-0.3cm}
\begin{figure}[htbp]
  \centering
  \includegraphics[width=\textwidth]{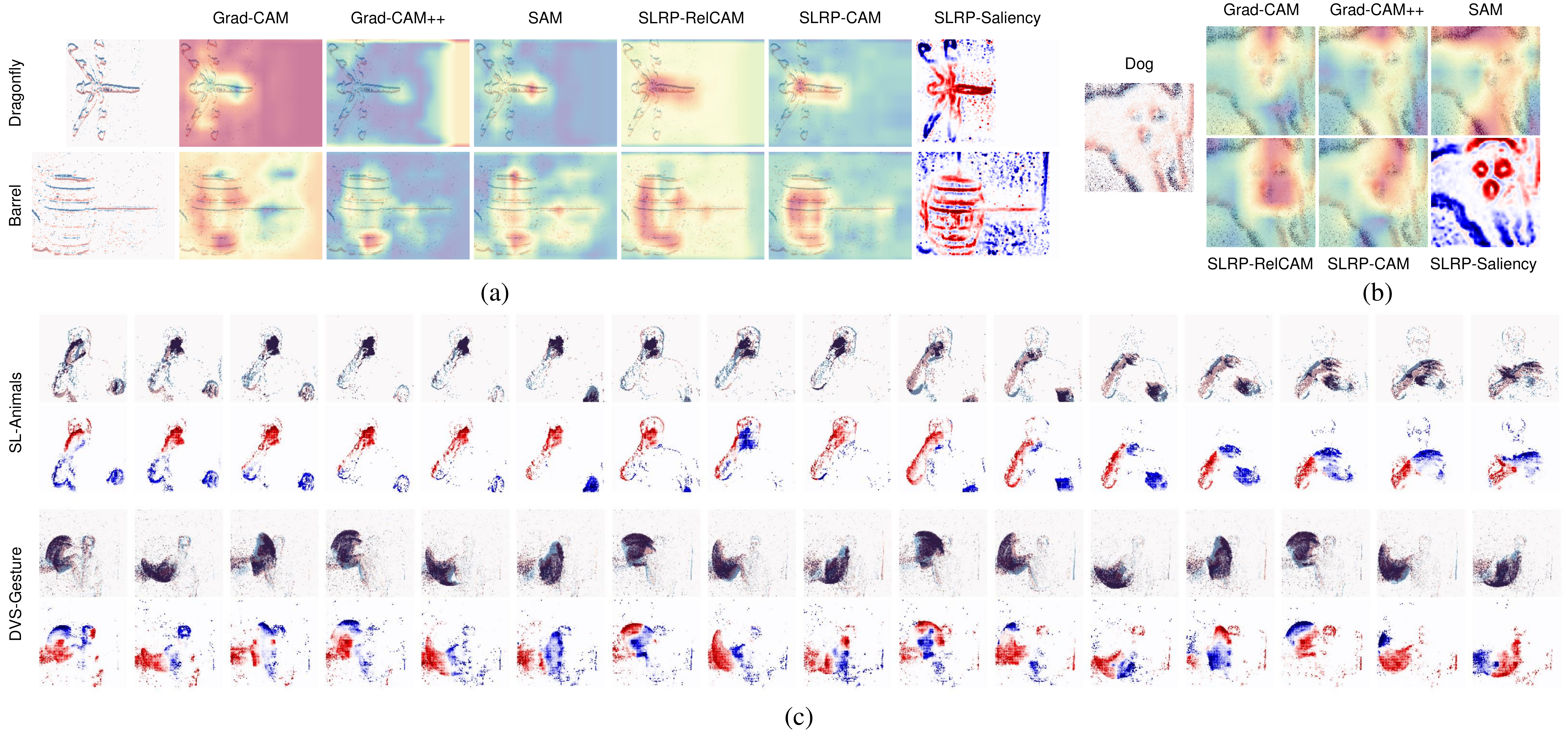}
  \caption{CAM and saliency map by different methods on \textbf{(a)} N-Caltech101 and \textbf{(b)} CIFAR10-DVS. \textbf{(c)} Saliency maps generated from \texttt{SLTRP} on DVS-Gesture and SL-Animals.}
  
  \label{fig:slrp}
  \vspace{-0.2cm}
  \end{figure}

\vspace{-0.3cm}
\subsubsection{Evaluation of Selectivity}
\vspace{-0.2cm}
We visualized the saliency map and CAM results of ours and other methods from Spiking-VGG11 and SEW Resnet18. As shown in \cref{fig:slrp}a, our method is more selective than other methods, with a higher value on the label-related objects and a lower value on the background. In contrast, Grad-CAM and SAM are more likely to be affected by the background, and Grad-CAM++ failed to locate the ``dragonfly'' on the top row, even though there are no other label-unralated events in this sample. On CIFAR10-DVS, SLRP-CAM and SLRP-RelCAM successfully localize the dog's head, whereas the attention of the other methods drifted to other regions. 

The saliency maps generated from \texttt{SLTRP} are able to track the exact moving object in the dataset (see \cref{fig:slrp}c). In the demonstration of SL-Animals dataset, the saliency map first focuses on the hand part to recognize the gesture. In the later time steps, it transfers its attention to the arm to capture the moving information. This proves its capability of temporal saliency information capturing , also yielding high selectivity.
\vspace{-0.3cm}
\begin{table}[h]
  \centering
  \resizebox{\textwidth}{!}{
    \begin{tabular}{cc|ccccccccc} 
      \toprule
      Model                           & \multicolumn{1}{c}{Resolution} & Grad-CAM & Grad-CAM++ & SLRP-RelCAM     & SLTRP-RelCAM    & SAM             & SLRP-CAM        & SLRP-Saliency Map & SLTRP-CAM       & SLTRP-Saliency Map  \\
      \midrule
      \multirow{2}{*}{Spiking VGG-11} & (48, 48)                       & 0.0776   & 0.0884     & \textbf{0.0225} & \textbf{0.0233} & \textbf{0.0267} & \textbf{0.0246} & 0.0645            & \textbf{0.0279} & 0.1157              \\
                                      & (128, 128)                     & 0.0842   & 0.0996     & \textbf{0.0335} & \textbf{0.0347} & \textbf{0.0303} & \textbf{0.0323} & 0.0716            & \textbf{0.0345} & 0.1886              \\
      \midrule
      SEW Resnet18                    & (128, 128)                     & 0.2629   & 0.2846     & \textbf{0.0902} & \textbf{0.1072} & \textbf{0.0909} & \textbf{0.0869} & 0.1960            & \textbf{0.0785} & 0.2881              \\
      \bottomrule
      \end{tabular}
  }
  \caption{Average time cost (s) of generating CAM and saliency map on N-Caltech101 (Spiking-VGG11) and DVSGesture (SEW Resnet18). Results within the quickest tier are highlighted in bold.}
  \label{tab:2}
  \vspace{-0.5cm}
\end{table}
\vspace{-0.2cm}
\subsubsection{Computation Time}
\vspace{-0.2cm}

As shown in \cref{tab:2}, SLRP-CAM, SLTRP-CAM are on the same level as SLTRP-RelCAM, SLRP-RelCAM, and SAM, all among the fastest methods. SLRP-Saliency Map and SLTRP-Saliency Map are slower than other methods, while still being competitive on SEW Resnet18 compared with Grad-CAM and Grad-CAM++. The time cost of SLTRP-CAM does not increase a lot compared to SLRP-CAM, since it only requires the relevance computation of the last fully connected layer. In contrast, the time cost of SLTRP-Saliency Map is much higher than SLRP-Saliency Map, since it requires the relevance computation of all layers.
\vspace{-0.2cm}
\subsection{Results of EventRPG}
\vspace{-0.2cm}
\begin{table}[htbp]
  \centering
  \resizebox{\textwidth}{!}{
    \begin{tabular}{cccccccc} 
      \toprule
      Dataset                                       & Data Augmentation                 & Training Method                               & Neural Network               & Neuron               & Timesteps           & Resolution                 & Accuracy        \\ 
      \midrule
      \multirow{8}{*}{\rotatebox{90}{N-Caltech101}} & Flip                              & SALT \citep{kim2021optimizing}                & Spike-VGG16                  & LIF                  & 20                  & (80,80)                    & 55.00           \\
                                                    & NDA \citep{NDA}                   & STBP-tdBN \citep{zheng2021going}              & Spike-VGG11                  & LIF                  & 10                  & (48,48)                    & 78.20           \\
                                                    & NDA \citep{NDA}                   & STBP-tdBN \citep{zheng2021going}              & Spike-VGG11                  & LIF                  & 10                  & (128,128)                  & 83.70           \\
                                                    & Eventmix \citep{shen2023eventmix} & STBP                                          & Pre-Act Resnet18             & PLIF                 & 10                  & (48, 48)                   & 79.47           \\ 
      \cmidrule{2-8}
                                                    & Identity                          & \multirow{4}{*}{TET \citep{deng2022temporal}} & \multirow{4}{*}{Spike-VGG11} & \multirow{4}{*}{LIF} & \multirow{4}{*}{10} & \multirow{4}{*}{(128,128)} & 75.70           \\
                                                    & EventDrop \citep{eventdrop}       &                                               &                              &                      &                     &                            & 74.04           \\
                                                    & EventRPG (CAM)                    &                                               &                              &                      &                     &                            & \uline{85.00}   \\
                                                    & EventRPG (Saliency Map)           &                                               &                              &                      &                     &                            & \textbf{85.62}  \\ 
      \midrule
      \multirow{12}{*}{\rotatebox{90}{CIFAR10-DVS}} & Drop by time                      & STBP                                          & SEW Wide-7B-Net              & PLIF                 & 16                  & (128,128)                  & 74.40           \\
                                                    & Flip                              & SALT \citep{kim2021optimizing}                & Spike-VGG16                  & LIF                  & 20                  & (64,64)                    & 67.10           \\
                                                    & Random Crop                       & DSR \citep{meng2022training}                  & Spike-VGG11                  & IF                   & 20                  & (48,48)                    & 75.03 ± 0.39    \\
                                                    & Random Crop                       & DSR \citep{meng2022training}                  & Spike-VGG11                  & LIF                  & 20                  & (48,48)                    & 77.27±0.24      \\
                                                    & FlipTranslation                   & TET \citep{deng2022temporal}                  & Spike-VGG11                  & LIF                  & 10                  & (48,48)                    & 83.17±0.15      \\
                                                    & NDA \citep{NDA}                   & STBP-tdBN \citep{zheng2021going}              & Spike-VGG11                  & LIF                  & 10                  & (48,48)                    & 79.60           \\
                                                    & NDA \citep{NDA}                   & STBP-tdBN \citep{zheng2021going}              & Spike-VGG11                  & LIF                  & 10                  & (128,128)                  & 81.70           \\
                                                    & Eventmix \citep{shen2023eventmix} & STBP                                          & Pre-Act Resnet18             & PLIF                 & 10                  & (48,48)                    & 81.45           \\ 
      \cmidrule{2-8}
                                                    & Identity                          & \multirow{4}{*}{TET \citep{deng2022temporal}} & \multirow{4}{*}{Spike-VGG11} & \multirow{4}{*}{LIF} & \multirow{4}{*}{10} & \multirow{4}{*}{(48, 48)}  & 78.85           \\
                                                    & EventDrop \citep{eventdrop}       &                                               &                              &                      &                     &                            & 77.73           \\
                                                    & EventRPG (CAM)                    &                                               &                              &                      &                     &                            & \textbf{85.55}  \\
                                                    & EventRPG (Saliency Map)           &                                               &                              &                      &                     &                            & \uline{84.96}   \\ 
      \midrule
      \multirow{7}{*}{\rotatebox{90}{N-Cars}}       & NDA \citep{NDA}                   & STBP-tdBN \citep{zheng2021going}              & Spike-VGG11                  & LIF                  & 10                  & (48,48)                    & 90.10           \\
                                                    & NDA \citep{NDA}                   & STBP-tdBN \citep{zheng2021going}              & Spike-VGG11                  & LIF                  & 10                  & (128,128)                  & 91.90           \\
                                                    & Eventmix \citep{shen2023eventmix} & STBP                                          & Pre-Act Resnet18             & PLIF                 & 10                  & (48,48)                    & \textbf{96.29}  \\ 
      \cmidrule{2-8}
                                                    & Identity                          & \multirow{4}{*}{TET \citep{deng2022temporal}} & \multirow{4}{*}{Spike-VGG11} & \multirow{4}{*}{LIF} & \multirow{4}{*}{10} & \multirow{4}{*}{(48,48)}   & 94.92           \\
                                                    & EventDrop \citep{eventdrop}       &                                               &                              &                      &                     &                            & 95.46           \\
                                                    & EventRPG (CAM)                    &                                               &                              &                      &                     &                            & 95.76           \\
                                                    & EventRPG (Saliency Map)           &                                               &                              &                      &                     &                            & \uline{96.00}   \\
      \bottomrule
      \end{tabular}
}
\caption{Accuracy of various data augmentation methods on event-based object recognition datasets. All of the datasets used are created using event cameras. Specifically, N-Caltech101 and CIFAR10DVS are derived from static images, while N-Cars is recorded in real-world environments.}
\vspace{-0.3cm}
\label{tab:3}
\end{table}

In this section, we evaluate our proposed \texttt{EventRPG} with other augmentation methods including EventDrop~\citep{eventdrop}, NDA~\citep{NDA}, and EventMix~\citep{shen2023eventmix} across several object recognition and action recognition datasets. They could illustrate the performance of our methods in terms of static objects and moving objects, respectively. The datasets used and training settings are introduced in \cref{app:dataset} in details. Since \texttt{SLRTP} and \texttt{SLRP} yield nearly identical results for object recognition tasks, we only leverage \texttt{SLRP} in object recognition experiments since it costs fewer time. Eventdrop~\citep{eventdrop} did not conduct experiments on SNNs, so we reproduce it using its public code and conduct experiments under the same training setting of ours. We report the best accuracy for each experiment using same random seed.
\vspace{-0.3cm}
\subsubsection{Object Recognition Tasks}
\vspace{-0.1cm}
From \cref{tab:3} we see that \texttt{EventRPG} achieves state-of-the-art performance on N-Caltech101 and CIFAR10-DVS datasets, bringing $9.92\%$ and $6.7\%$ improvements compared with identity (no augmentation), respectively. On N-Cars, \texttt{EventRPG} achieves the second-best performance, only $0.29\%$ lower than EventMix. This might be attributed to the fact that N-Cars is a binary classification dataset, which only contains label ``car'' and ``background''. Most samples belonging to ``background'' do not have a specific label-related object to locate, making it difficult for our method to generate saliency map and CAM with high quality, thus decreasing the performance.

\begin{table}
  \centering
  \resizebox{0.8\textwidth}{!}{
    \begin{tabular}{cccccc} 
      \toprule
      \multirow{2}{*}{Model}        & \multirow{2}{*}{Spike}      & \multirow{2}{*}{Method}           & \multirow{2}{*}{DVSGesture} & \multicolumn{2}{c}{SL-Animals}   \\ 
      \cmidrule{5-6}
                                    &                             &                                   &                             & 4 sets         & 3 sets          \\ 
      \midrule
      7-Layer Spiking CNN           & Hybrid                      & SCTFA \citep{cai2023spatial}       & \textbf{98.96}              & \textbf{90.04} & -               \\
      GoogLeNet                     & \XSolidBrush                & TORE \citep{baldwin2022time}      & 96.20                        & 85.10          & -               \\
      Event Transformer             & \XSolidBrush                & EvT \citep{sabater2022event}      & 96.20                       & \uline{88.12}  & \textbf{87.45}  \\
      Pre-Act Resnet18              & \Checkmark                  & EventMix \citep{shen2023eventmix} & \uline{96.75}               & -              & -               \\ 
      \midrule
      \multirow{7}{*}{SEW Resnet18} & \multirow{7}{*}{\Checkmark} & Identity                          & 94.33                       & 85.42          & 89.09           \\
                                    &                             & EventDrop \citep{eventdrop}       & 92.33                       & 86.33          & 88.99           \\
                                    &                             & NDA \citep{NDA}                   & 93.67                       & 87.77          & 89.55           \\
                                    &                             & EventRPG (CAM, SLRP)              & 95.83                       & \uline{90.97}  & 91.96           \\
                                    &                             & EventRPG (Saliency Map, SLRP)     & 95.49                       & \textbf{91.59} & \uline{93.30}    \\
                                    &                             & EventRPG (CAM, SLTRP)             & \uline{96.18}               & 90.54          & 90.63           \\
                                    &                             & EventRPG (Saliency Map, SLTRP)    & \textbf{96.53}              & 90.04          & \textbf{93.75}  \\
      \bottomrule
      \end{tabular}
  }
  \caption{Accuracy of various data augmentation methods on event-based action recognition datasets.}
  \label{tab:4}
  \vspace{-0.6cm}
  \end{table}

  \vspace{-0.2cm}
\subsubsection{Action Recognition Tasks}
\vspace{-0.1cm}
We implement \texttt{EventRPG} on SEW Resnet18 for action recognition tasks. On DVSGesture dataset, our method achieves best results compared with other reproduced augmentation methods under the same training settings, though slightly lower than EventMix on Pre-Act Resnet18. On SL-Animals dataset, our method achieves state-of-the-art performance among all data augmentation apporaches and all neural networks, with $3.82\%$ and $4.2\%$ improvements compared with the second-best augmentation method on 4 sets and 3 sets, respectively. Action recognition tasks include more dynamic information compared to object recognition tasks. Thus, the success of our method on action recognition tasks demonstrates its great potential for other dynamic event-based datasets, which is a future direction worth exploring.
\vspace{-0.2cm}
\subsubsection{Time Consumption Analysis}
\vspace{-0.4cm}
\begin{figure}[htbp]
  \centering  
    \begin{subfigure}{0.49\textwidth}
      \centering   
      \includegraphics[width=\linewidth]{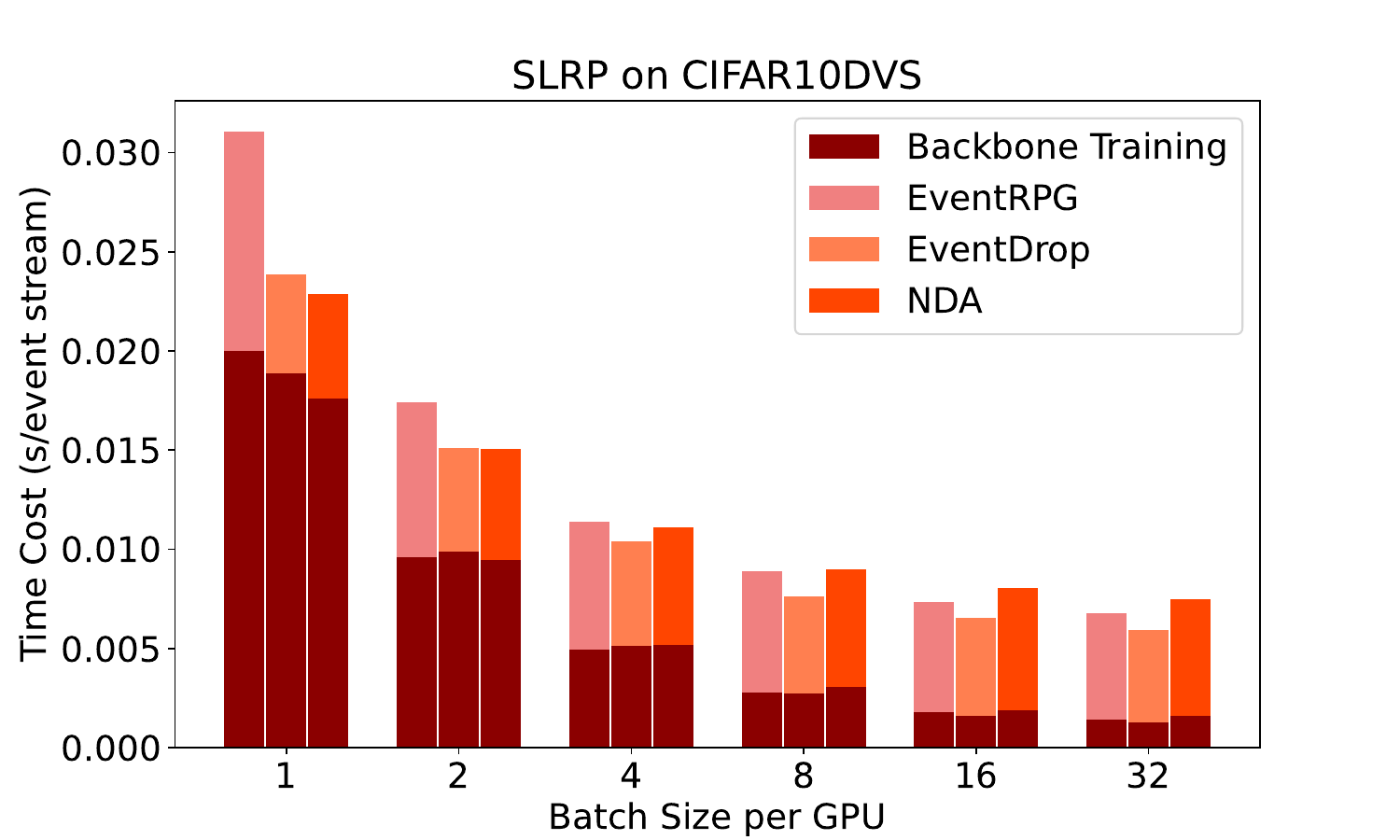}
        \caption{}
    \end{subfigure}
    \begin{subfigure}{0.49\textwidth}
      \centering   
      \includegraphics[width=\linewidth]{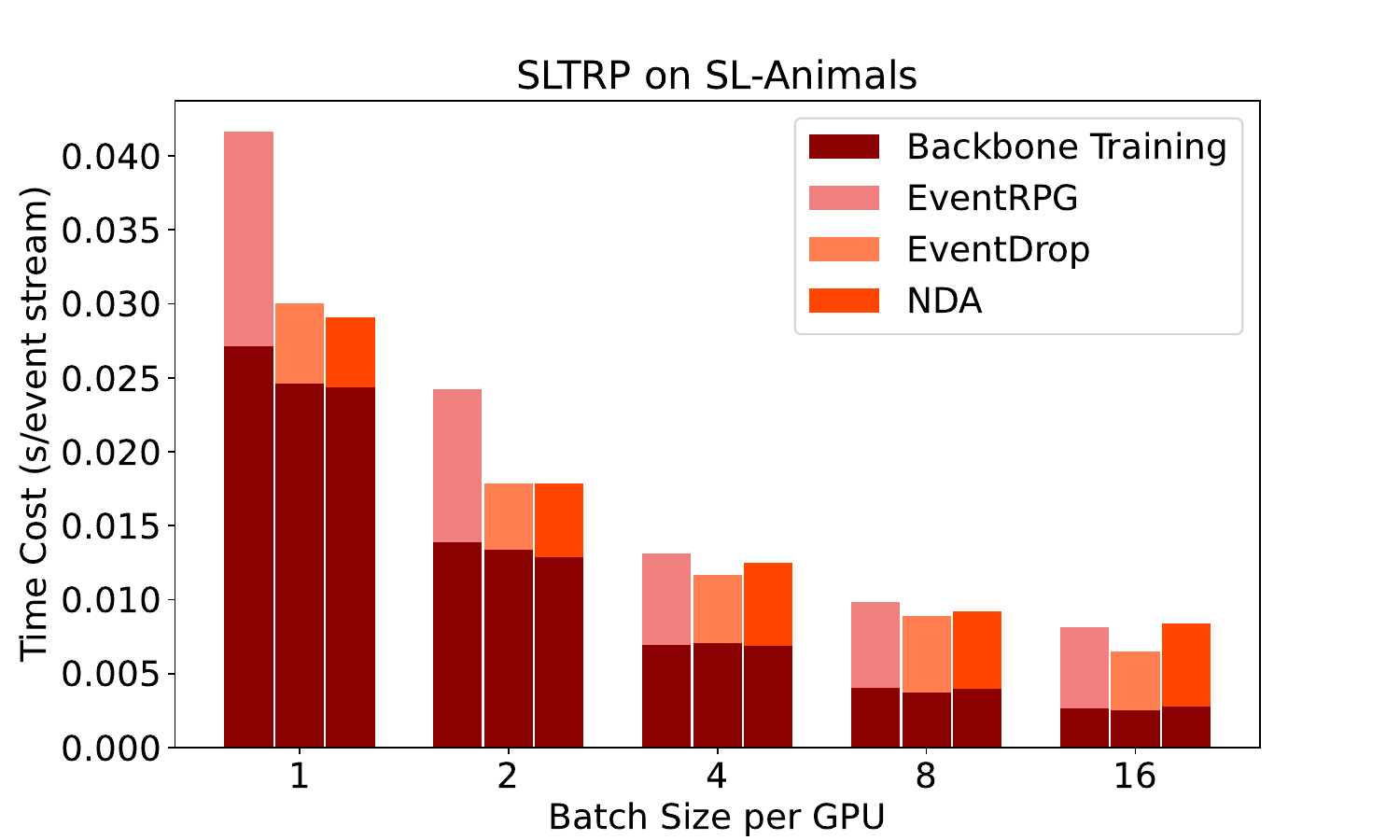}
        \caption{}
    \end{subfigure}
\caption{
\label{fig:timecost}
Average time cost of augmentation methods with SEW Resnet18 on two datasets.
}
\end{figure}
\vspace{-0.3cm}
In order to assess the computational efficiency of our data augmentation method, we perform two experiments to measure the time required for executing \texttt{EventRPG} with \texttt{SLRP} and \texttt{SLTRP} in object recognition and action recognition tasks. As depicted in \cref{fig:timecost}, the computation time of other data augmentation approaches remains constant regardless of the batch size, whereas our method exhibits a nearly linear decrease in computation time as the batch size increases, since the relevance propagation process, which is the main contributor to the time consumption, can be speeded up by parallizing computing across samples in a batch, similar to the gradient backpropagation process. When the batch size for each GPU exceeds 4, both \texttt{SLRP} and \texttt{SLTRP} achieve comparable speeds to NDA and EventDrop, confirming the time efficiency of \texttt{EventRPG}.
\vspace{-0.3cm}
\section{Conclusion and Limitation}
\vspace{-0.3cm}
\paragraph{Conclusion} In this paper, for the first time, we propose \texttt{SLTRP} and \texttt{SLRP}, two efficient and practical methods for generating CAMs and saliency maps for SNNs. Building upon this, we propose \texttt{EventRPG}, i.e., dropping events and mixing events with Relevance Propagation Guidance. Since \texttt{EventRPG} only disturbs and mixes regions on which model concerns most, it is more efficient compared to vanilla dropping and mixing, and also alleviates the likely misalignment problem between data and label. In our experiments, \texttt{SLRP} and \texttt{SLTRP} not only both yeild best results compared with other feature visualization tools, but also consume very little time to compute. \texttt{EventRPG} achieves state-of-the-art performance on N-Caltech101, CIFAR10-DVS and SL-Animals datasets, proving its strong generalization ability across different models and datasets. 
\vspace{-0.3cm}
\paragraph{Limitation} Currently, \texttt{EventRPG} can only be implemented on classification tasks, remaining as a limitation. However, we could still leverage multi-task training paradim which has been proven to be effective to implement it into other downstream tasks, and we will also explore more possibilities of using \texttt{EventRPG} in self-supervised learning tasks.

\section{Acknowledgement}
We would like to thank all anonymous reviewers for their committed work and insightful feedback. This work was supported in part by the National Natural Science Foundation of China under Grants 62073066 and U20A20197, in part by the Fundamental Research Funds for the Central Universities under Grant N2226001, in part by 111 Project under Grant B16009, in part by the Intel Neuromorphic Research Community (INRC) Grant Award (RV2.137.Fang), and in part by Guangzhou-HKUST(GZ) Joint Funding Program under Grant 2023A03J0682.
\bibliographystyle{iclr2024_conference}
\bibliography{EventRPG}

\begin{thebibliography}{53}
\providecommand{\natexlab}[1]{#1}
\providecommand{\url}[1]{\texttt{#1}}
\expandafter\ifx\csname urlstyle\endcsname\relax
  \providecommand{\doi}[1]{doi: #1}\else
  \providecommand{\doi}{doi: \begingroup \urlstyle{rm}\Url}\fi

\bibitem[Amir et~al.(2017)Amir, Taba, Berg, Melano, McKinstry, Di~Nolfo, Nayak,
  Andreopoulos, Garreau, Mendoza, Kusnitz, Debole, Esser, Delbruck, Flickner,
  and Modha]{dvs_gesture}
Arnon Amir, Brian Taba, David Berg, Timothy Melano, Jeffrey McKinstry, Carmelo
  Di~Nolfo, Tapan Nayak, Alexander Andreopoulos, Guillaume Garreau, Marcela
  Mendoza, Jeff Kusnitz, Michael Debole, Steve Esser, Tobi Delbruck, Myron
  Flickner, and Dharmendra Modha.
\newblock A low power, fully event-based gesture recognition system.
\newblock In \emph{CVPR}, July 2017.

\bibitem[Bach et~al.(2015)Bach, Binder, Montavon, Klauschen, M{\"u}ller, and
  Samek]{bach2015pixel}
Sebastian Bach, Alexander Binder, Gr{\'e}goire Montavon, Frederick Klauschen,
  Klaus-Robert M{\"u}ller, and Wojciech Samek.
\newblock {On Pixel-wise Explanations for Non-linear Classifier Decisions by
  Layer-wise Relevance Propagation}.
\newblock \emph{{PloS One}}, 10\penalty0 (7):\penalty0 e0130140, 2015.

\bibitem[Baldwin et~al.(2022)Baldwin, Liu, Almatrafi, Asari, and
  Hirakawa]{baldwin2022time}
R~Wes Baldwin, Ruixu Liu, Mohammed Almatrafi, Vijayan Asari, and Keigo
  Hirakawa.
\newblock Time-ordered recent event (tore) volumes for event cameras.
\newblock \emph{IEEE TPAMI}, 45\penalty0 (2):\penalty0 2519--2532, 2022.

\bibitem[Bardow et~al.(2016)Bardow, Davison, and
  Leutenegger]{bardow2016simultaneous}
Patrick Bardow, Andrew~J Davison, and Stefan Leutenegger.
\newblock {Simultaneous Optical Flow and Intensity Estimation from An Event
  Camera}.
\newblock In \emph{CVPR}, pp.\  884--892, 2016.

\bibitem[Baudron et~al.(2020)Baudron, Wang, Cossairt, and
  Katsaggelos]{baudron2020e3d}
Alexis Baudron, Zihao~W Wang, Oliver Cossairt, and Aggelos~K Katsaggelos.
\newblock {E3D: Event-Based 3D Shape Reconstruction}.
\newblock \emph{arXiv preprint arXiv:2012.05214}, 2020.

\bibitem[Cai et~al.(2023)Cai, Sun, Liu, Cui, Wang, Xia, Yao, and
  Guo]{cai2023spatial}
Wuque Cai, Hongze Sun, Rui Liu, Yan Cui, Jun Wang, Yang Xia, Dezhong Yao, and
  Daqing Guo.
\newblock A spatial--channel--temporal-fused attention for spiking neural
  networks.
\newblock \emph{TNNLS}, 2023.

\bibitem[Chattopadhay et~al.(2018)Chattopadhay, Sarkar, Howlader, and
  Balasubramanian]{chattopadhay2018grad}
Aditya Chattopadhay, Anirban Sarkar, Prantik Howlader, and Vineeth~N
  Balasubramanian.
\newblock {Grad-cam++: Generalized Gradient-based Visual Explanations for Deep
  Convolutional Networks}.
\newblock In \emph{WACV}, pp.\  839--847. IEEE, 2018.

\bibitem[Chen et~al.(2023)Chen, Guan, and Lu]{chen2023esvio}
Peiyu Chen, Weipeng Guan, and Peng Lu.
\newblock Esvio: Event-based stereo visual inertial odometry.
\newblock \emph{IEEE Robotics and Automation Letters}, 2023.

\bibitem[Deng et~al.(2022)Deng, Li, Zhang, and Gu]{deng2022temporal}
Shikuang Deng, Yuhang Li, Shanghang Zhang, and Shi Gu.
\newblock Temporal efficient training of spiking neural network via gradient
  re-weighting.
\newblock In \emph{ICLR}, 2022.

\bibitem[Fang et~al.(2020)Fang, Chen, Ding, Chen, Yu, Zhou, Tian, and other
  contributors]{SpikingJelly}
Wei Fang, Yanqi Chen, Jianhao Ding, Ding Chen, Zhaofei Yu, Huihui Zhou,
  Yonghong Tian, and other contributors.
\newblock Spikingjelly.
\newblock \url{https://github.com/fangwei123456/spikingjelly}, 2020.
\newblock Accessed: 2022-09-19.

\bibitem[Fang et~al.(2021{\natexlab{a}})Fang, Yu, Chen, Huang, Masquelier, and
  Tian]{fang2021deep}
Wei Fang, Zhaofei Yu, Yanqi Chen, Tiejun Huang, Timoth{\'e}e Masquelier, and
  Yonghong Tian.
\newblock Deep residual learning in spiking neural networks.
\newblock \emph{NeurIPS}, 34:\penalty0 21056--21069, 2021{\natexlab{a}}.

\bibitem[Fang et~al.(2021{\natexlab{b}})Fang, Yu, Chen, Masquelier, Huang, and
  Tian]{fang2021incorporating}
Wei Fang, Zhaofei Yu, Yanqi Chen, Timoth{\'e}e Masquelier, Tiejun Huang, and
  Yonghong Tian.
\newblock Incorporating learnable membrane time constant to enhance learning of
  spiking neural networks.
\newblock In \emph{ICCV}, pp.\  2661--2671, 2021{\natexlab{b}}.

\bibitem[Fei-Fei et~al.(2004)Fei-Fei, Fergus, and Perona]{fei2004learning}
Li~Fei-Fei, Rob Fergus, and Pietro Perona.
\newblock {Learning Generative Visual Models from Few Training Examples: An
  Incremental Bayesian Approach Tested on 101 Object Categories}.
\newblock In \emph{CVPRW}, pp.\  178--178. IEEE, 2004.

\bibitem[Gallego et~al.(2020)Gallego, Delbr{\"u}ck, Orchard, Bartolozzi, Taba,
  Censi, Leutenegger, Davison, Conradt, Daniilidis,
  et~al.]{gallego2020eventsurvey}
Guillermo Gallego, Tobi Delbr{\"u}ck, Garrick Orchard, Chiara Bartolozzi, Brian
  Taba, Andrea Censi, Stefan Leutenegger, Andrew~J Davison, J{\"o}rg Conradt,
  Kostas Daniilidis, et~al.
\newblock {Event-based Vision: A Survey}.
\newblock \emph{IEEE TPAMI}, 44\penalty0 (1):\penalty0 154--180, 2020.

\bibitem[Gehrig et~al.(2019)Gehrig, Loquercio, Derpanis, and
  Scaramuzza]{gehrig2019end}
Daniel Gehrig, Antonio Loquercio, Konstantinos~G Derpanis, and Davide
  Scaramuzza.
\newblock {End-to-end Learning of Representations for Asynchronous Event-based
  Data}.
\newblock In \emph{ICCV}, pp.\  5633--5643, 2019.

\bibitem[Gehrig et~al.(2021)Gehrig, Millh{\"a}usler, Gehrig, and
  Scaramuzza]{gehrig2021raft}
Mathias Gehrig, Mario Millh{\"a}usler, Daniel Gehrig, and Davide Scaramuzza.
\newblock {E-raft: Dense Optical Flow from Event Cameras}.
\newblock In \emph{3DV}, pp.\  197--206. IEEE, 2021.

\bibitem[Gu et~al.(2021)Gu, Sng, Hu, and Yu]{eventdrop}
Fuqiang Gu, Weicong Sng, Xuke Hu, and Fangwen Yu.
\newblock {EventDrop: Data Augmentation for Event-based Learning}.
\newblock In \emph{IJCAI}, 2021.
\newblock URL \url{https://arxiv.org/abs/2106.05836}.

\bibitem[Gu et~al.(2018)Gu, Yang, and Tresp]{gu2018understanding}
Jindong Gu, Yinchong Yang, and Volker Tresp.
\newblock Understanding individual decisions of cnns via contrastive
  backpropagation.
\newblock In \emph{ACCV}, pp.\  119--134. Springer, 2018.

\bibitem[Hendrycks et~al.(2019)Hendrycks, Mu, Cubuk, Zoph, Gilmer, and
  Lakshminarayanan]{hendrycks2019augmix}
Dan Hendrycks, Norman Mu, Ekin~D Cubuk, Barret Zoph, Justin Gilmer, and Balaji
  Lakshminarayanan.
\newblock {Augmix: A Simple Data Processing Method to Improve Robustness and
  Uncertainty}.
\newblock In \emph{ICLR}, 2019.

\bibitem[Kim et~al.(2020)Kim, Choo, and Song]{kim2020puzzle}
Jang-Hyun Kim, Wonho Choo, and Hyun~Oh Song.
\newblock {Puzzle Mix: Exploiting Saliency and Local Statistics for Optimal
  Mixup}.
\newblock In \emph{ICML}, 2020.

\bibitem[Kim et~al.(2021)Kim, Bae, Park, Zhang, and Kim]{kim2021n}
Junho Kim, Jaehyeok Bae, Gangin Park, Dongsu Zhang, and Young~Min Kim.
\newblock N-imagenet: Towards robust, fine-grained object recognition with
  event cameras.
\newblock In \emph{ICCV}, pp.\  2146--2156, 2021.

\bibitem[Kim \& Panda(2021{\natexlab{a}})Kim and Panda]{kim2021optimizing}
Youngeun Kim and Priyadarshini Panda.
\newblock Optimizing deeper spiking neural networks for dynamic vision sensing.
\newblock \emph{Neural Networks}, 144:\penalty0 686--698, 2021{\natexlab{a}}.

\bibitem[Kim \& Panda(2021{\natexlab{b}})Kim and Panda]{kim2021visual}
Youngeun Kim and Priyadarshini Panda.
\newblock Visual explanations from spiking neural networks using inter-spike
  intervals.
\newblock \emph{Scientific reports}, 11\penalty0 (1):\penalty0 1--14,
  2021{\natexlab{b}}.

\bibitem[Kim et~al.(2022)Kim, Joshua, and Panda]{kim2022beyond}
Youngeun Kim, Chough Joshua, and Priyadarshini Panda.
\newblock Beyond classification: Directly training spiking neural networks for
  semantic segmentation.
\newblock \emph{Neuromorphic Computing and Engineering}, 2022.

\bibitem[Kogler et~al.(2009)Kogler, Sulzbachner, and Kubinger]{kogler2009bio}
J{\"u}rgen Kogler, Christoph Sulzbachner, and Wilfried Kubinger.
\newblock {Bio-inspired Stereo Vision System with Silicon Retina Imagers}.
\newblock In \emph{International Conference on Computer Vision Systems}, pp.\
  174--183. Springer, 2009.

\bibitem[Krizhevsky et~al.(2009)Krizhevsky, Hinton,
  et~al.]{krizhevsky2009learning}
Alex Krizhevsky, Geoffrey Hinton, et~al.
\newblock Learning multiple layers of features from tiny images.
\newblock 2009.

\bibitem[Lagorce et~al.(2016)Lagorce, Orchard, Galluppi, Shi, and
  Benosman]{lagorce2016hots}
Xavier Lagorce, Garrick Orchard, Francesco Galluppi, Bertram~E Shi, and Ryad~B
  Benosman.
\newblock Hots: a hierarchy of event-based time-surfaces for pattern
  recognition.
\newblock \emph{IEEE TPAMI}, 39\penalty0 (7):\penalty0 1346--1359, 2016.

\bibitem[Lee et~al.(2020)Lee, Kosta, Zhu, Chaney, Daniilidis, and
  Roy]{lee2020spike}
Chankyu Lee, Adarsh~Kumar Kosta, Alex~Zihao Zhu, Kenneth Chaney, Kostas
  Daniilidis, and Kaushik Roy.
\newblock {Spike-flownet: Event-based Optical Flow Estimation with
  Energy-efficient Hybrid Neural Networks}.
\newblock In \emph{ECCV}, pp.\  366--382. Springer, 2020.

\bibitem[Lee et~al.(2021)Lee, Kim, Park, Eo, and Hwang]{lee2021relevance}
Jeong~Ryong Lee, Sewon Kim, Inyong Park, Taejoon Eo, and Dosik Hwang.
\newblock {Relevance-cam: Your Model Already Knows Where to Look}.
\newblock In \emph{CVPR}, pp.\  14944--14953, 2021.

\bibitem[Li et~al.(2017)Li, Liu, Ji, Li, and Shi]{li2017cifar10}
Hongmin Li, Hanchao Liu, Xiangyang Ji, Guoqi Li, and Luping Shi.
\newblock {Cifar10-dvs: an Event-stream Dataset for Object Classification}.
\newblock \emph{Frontiers in Neuroscience}, 11:\penalty0 309, 2017.

\bibitem[Li et~al.(2022)Li, Kim, Park, Geller, and Panda]{NDA}
Yuhang Li, Youngeun Kim, Hyoungseob Park, Tamar Geller, and Priyadarshini
  Panda.
\newblock {Neuromorphic Data Augmentation for Training Spiking Neural
  Networks}.
\newblock In \emph{ECCV}, 2022.
\newblock URL \url{https://arxiv.org/abs/2203.06145}.

\bibitem[Maass(1997)]{maass1997networks}
Wolfgang Maass.
\newblock {Networks of Spiking Neurons: the Third Generation of Neural Network
  Models}.
\newblock \emph{Neural Networks}, 10\penalty0 (9):\penalty0 1659--1671, 1997.

\bibitem[Mahowald(1994)]{mahowald1994silicon}
Misha Mahowald.
\newblock {The Silicon Retina}.
\newblock In \emph{An Analog VLSI System for Stereoscopic Vision}, pp.\  4--65.
  Springer, 1994.

\bibitem[Meng et~al.(2022)Meng, Xiao, Yan, Wang, Lin, and
  Luo]{meng2022training}
Qingyan Meng, Mingqing Xiao, Shen Yan, Yisen Wang, Zhouchen Lin, and Zhi-Quan
  Luo.
\newblock {Training High-Performance Low-Latency Spiking Neural Networks by
  Differentiation on Spike Representation}.
\newblock In \emph{CVPR}, pp.\  12444--12453, 2022.

\bibitem[Montavon et~al.(2017)Montavon, Lapuschkin, Binder, Samek, and
  M{\"u}ller]{montavon2017explaining}
Gr{\'e}goire Montavon, Sebastian Lapuschkin, Alexander Binder, Wojciech Samek,
  and Klaus-Robert M{\"u}ller.
\newblock {Explaining Nonlinear Classification Decisions with Deep Taylor
  Decomposition}.
\newblock \emph{Pattern Recognition}, 65:\penalty0 211--222, 2017.

\bibitem[Orchard et~al.(2015)Orchard, Jayawant, Cohen, and
  Thakor]{orchard2015converting}
Garrick Orchard, Ajinkya Jayawant, Gregory~K Cohen, and Nitish Thakor.
\newblock {Converting Static Image Datasets to Spiking Neuromorphic Datasets
  using Saccades}.
\newblock \emph{Frontiers in Neuroscience}, 9:\penalty0 437, 2015.

\bibitem[Rudnev et~al.(2022)Rudnev, Elgharib, Theobalt, and
  Golyanik]{rudnev2022eventnerf}
Viktor Rudnev, Mohamed Elgharib, Christian Theobalt, and Vladislav Golyanik.
\newblock {EventNeRF: Neural Radiance Fields from a Single Colour Event
  Camera}.
\newblock \emph{arXiv preprint arXiv:2206.11896}, 2022.

\bibitem[Sabater et~al.(2022)Sabater, Montesano, and Murillo]{sabater2022event}
Alberto Sabater, Luis Montesano, and Ana~C Murillo.
\newblock Event transformer. a sparse-aware solution for efficient event data
  processing.
\newblock In \emph{CVPR}, pp.\  2677--2686, 2022.

\bibitem[Selvaraju et~al.(2017)Selvaraju, Cogswell, Das, Vedantam, Parikh, and
  Batra]{selvaraju2017grad}
Ramprasaath~R Selvaraju, Michael Cogswell, Abhishek Das, Ramakrishna Vedantam,
  Devi Parikh, and Dhruv Batra.
\newblock {Grad-cam: Visual Explanations from Deep Networks via Gradient-based
  Localization}.
\newblock In \emph{ICCV}, pp.\  618--626, 2017.

\bibitem[Shen et~al.(2023)Shen, Zhao, and Zeng]{shen2023eventmix}
Guobin Shen, Dongcheng Zhao, and Yi~Zeng.
\newblock Eventmix: An efficient data augmentation strategy for event-based
  learning.
\newblock \emph{Information Sciences}, 644:\penalty0 119170, 2023.

\bibitem[Sironi et~al.(2018)Sironi, Brambilla, Bourdis, Lagorce, and
  Benosman]{sironi2018hats}
Amos Sironi, Manuele Brambilla, Nicolas Bourdis, Xavier Lagorce, and Ryad
  Benosman.
\newblock Hats: Histograms of averaged time surfaces for robust event-based
  object classification.
\newblock In \emph{CVPR}, pp.\  1731--1740, 2018.

\bibitem[Son et~al.(2017)Son, Suh, Kim, Jung, Kim, Shin, Park, Lee, Park, Woo,
  et~al.]{son20174}
Bongki Son, Yunjae Suh, Sungho Kim, Heejae Jung, Jun-Seok Kim, Changwoo Shin,
  Keunju Park, Kyoobin Lee, Jinman Park, Jooyeon Woo, et~al.
\newblock {4.1 A 640$\times$ 480 Dynamic Vision Sensor with a 9$\mu$m Pixel and
  300Meps Address-event Representation}.
\newblock In \emph{ISSCC}, pp.\  66--67. IEEE, 2017.

\bibitem[Stoffregen et~al.(2019)Stoffregen, Gallego, Drummond, Kleeman, and
  Scaramuzza]{stoffregen2019event}
Timo Stoffregen, Guillermo Gallego, Tom Drummond, Lindsay Kleeman, and Davide
  Scaramuzza.
\newblock {Event-based Motion Segmentation by Motion Compensation}.
\newblock In \emph{ICCV}, pp.\  7244--7253, 2019.

\bibitem[Uddin et~al.(2020)Uddin, Monira, Shin, Chung, and
  Bae]{uddin2020saliencymix}
AFM~Shahab Uddin, Mst~Sirazam Monira, Wheemyung Shin, TaeChoong Chung, and
  Sung-Ho Bae.
\newblock {SaliencyMix: A Saliency Guided Data Augmentation Strategy for Better
  Regularization}.
\newblock In \emph{ICLR}, 2020.

\bibitem[Vasudevan et~al.(2021)Vasudevan, Negri, Di~Ielsi, Linares-Barranco,
  and Serrano-Gotarredona]{vasudevan2021sl}
Ajay Vasudevan, Pablo Negri, Camila Di~Ielsi, Bernabe Linares-Barranco, and
  Teresa Serrano-Gotarredona.
\newblock Sl-animals-dvs: event-driven sign language animals dataset.
\newblock \emph{Pattern Analysis and Applications}, pp.\  1--16, 2021.

\bibitem[Wang et~al.(2020)Wang, Wang, Du, Yang, Zhang, Ding, Mardziel, and
  Hu]{wang2020score}
Haofan Wang, Zifan Wang, Mengnan Du, Fan Yang, Zijian Zhang, Sirui Ding, Piotr
  Mardziel, and Xia Hu.
\newblock {Score-CAM: Score-weighted Visual Explanations for Convolutional
  Neural Networks}.
\newblock In \emph{CVPRW}, pp.\  24--25, 2020.

\bibitem[Wang et~al.(2021)Wang, Pan, Ng, Zhuang, and Mahony]{wang2021stereo}
Ziwei Wang, Liyuan Pan, Yonhon Ng, Zheyu Zhuang, and Robert Mahony.
\newblock Stereo hybrid event-frame (shef) cameras for 3d perception.
\newblock In \emph{IROS}, pp.\  9758--9764. IEEE, 2021.

\bibitem[Yun et~al.(2019)Yun, Han, Oh, Chun, Choe, and Yoo]{yun2019cutmix}
Sangdoo Yun, Dongyoon Han, Seong~Joon Oh, Sanghyuk Chun, Junsuk Choe, and
  Youngjoon Yoo.
\newblock {Cutmix: Regularization Strategy to Train Strong Classifiers with
  Localizable Features}.
\newblock In \emph{ICCV}, pp.\  6023--6032, 2019.

\bibitem[Zhang et~al.(2018)Zhang, Cisse, Dauphin, and
  Lopez-Paz]{zhang2018mixup}
Hongyi Zhang, Moustapha Cisse, Yann~N Dauphin, and David Lopez-Paz.
\newblock {Mixup: Beyond Empirical Risk Minimization}.
\newblock In \emph{ICLR}, 2018.

\bibitem[Zheng et~al.(2021)Zheng, Wu, Deng, Hu, and Li]{zheng2021going}
Hanle Zheng, Yujie Wu, Lei Deng, Yifan Hu, and Guoqi Li.
\newblock Going deeper with directly-trained larger spiking neural networks.
\newblock In \emph{AAAI}, volume~35, pp.\  11062--11070, 2021.

\bibitem[Zhou et~al.(2016)Zhou, Khosla, Lapedriza, Oliva, and
  Torralba]{zhou2016learning}
Bolei Zhou, Aditya Khosla, Agata Lapedriza, Aude Oliva, and Antonio Torralba.
\newblock {Learning Deep Features for Discriminative Localization}.
\newblock In \emph{CVPR}, pp.\  2921--2929, 2016.

\bibitem[Zhou et~al.(2021)Zhou, Gallego, Lu, Liu, and Shen]{zhou2021event}
Yi~Zhou, Guillermo Gallego, Xiuyuan Lu, Siqi Liu, and Shaojie Shen.
\newblock {Event-based Motion Segmentation with Spatio-temporal Graph Cuts}.
\newblock \emph{TNNLS}, 2021.

\bibitem[Zuo et~al.(2022)Zuo, Yang, Chen, Wang, Wang, and Kneip]{devo}
Yi-Fan Zuo, Jiaqi Yang, Jiaben Chen, Xia Wang, Yifu Wang, and Laurent Kneip.
\newblock {DEVO: Depth-Event Camera Visual Odometry in Challenging Conditions}.
\newblock \emph{ICRA}, 2022.

\end{thebibliography}

\clearpage
\appendix
{\LARGE{\textsc{Appendix}}}
\section{Additional Experimental Results}
Here we present more experiments to further prove the effectiveness of our approach.
\subsection{Classification on mini N-ImageNet Dataset}
N-ImageNet, as proposed by \citet{kim2021n}, represents the neuromorphic adaptation of the well-known ImageNet dataset. It encompasses a thousand object categories, making it the most challenging task in event-based classification to date. We conduct experients to evaluate the performances of different data augmentation methods on mini N-ImageNet with SEW Resnet-18.

\begin{table}[h]
  \centering
  \resizebox{0.6\textwidth}{!}{
  \begin{tabular}{ccccc} 
  \toprule
  Data Augmentation & Identity & EventDrop & NDA   & EventRPG  \\ 
  \midrule
  Top-1 Accuracy      & 28.16    & 34.18     & 35.84 & \textbf{40.90}     \\
  Top-5 Accuracy      & 52.14    & 60.94     & 63.64 & \textbf{67.74}     \\
  \bottomrule
  \end{tabular}
  }
  \caption{Top-1 and Top-5 Accuracies (\%) on mini N-ImageNet.}
  \label{tab:miniNImagenet}
\end{table}

\Cref{tab:miniNImagenet} demonstrates that data augmentations significantly enhance the performance of the model, as shown by the notable positive effects. In this comparison, our approach outperforms the nearest competitor NDA, achieving a higher Top-1 accuracy by $5.06\%$ and Top-5 accuracy by $4.1\%$.
\begin{figure}[htbp]
  \centering  
    \begin{subfigure}{0.49\textwidth}
      \centering   
      \includegraphics[width=\linewidth]{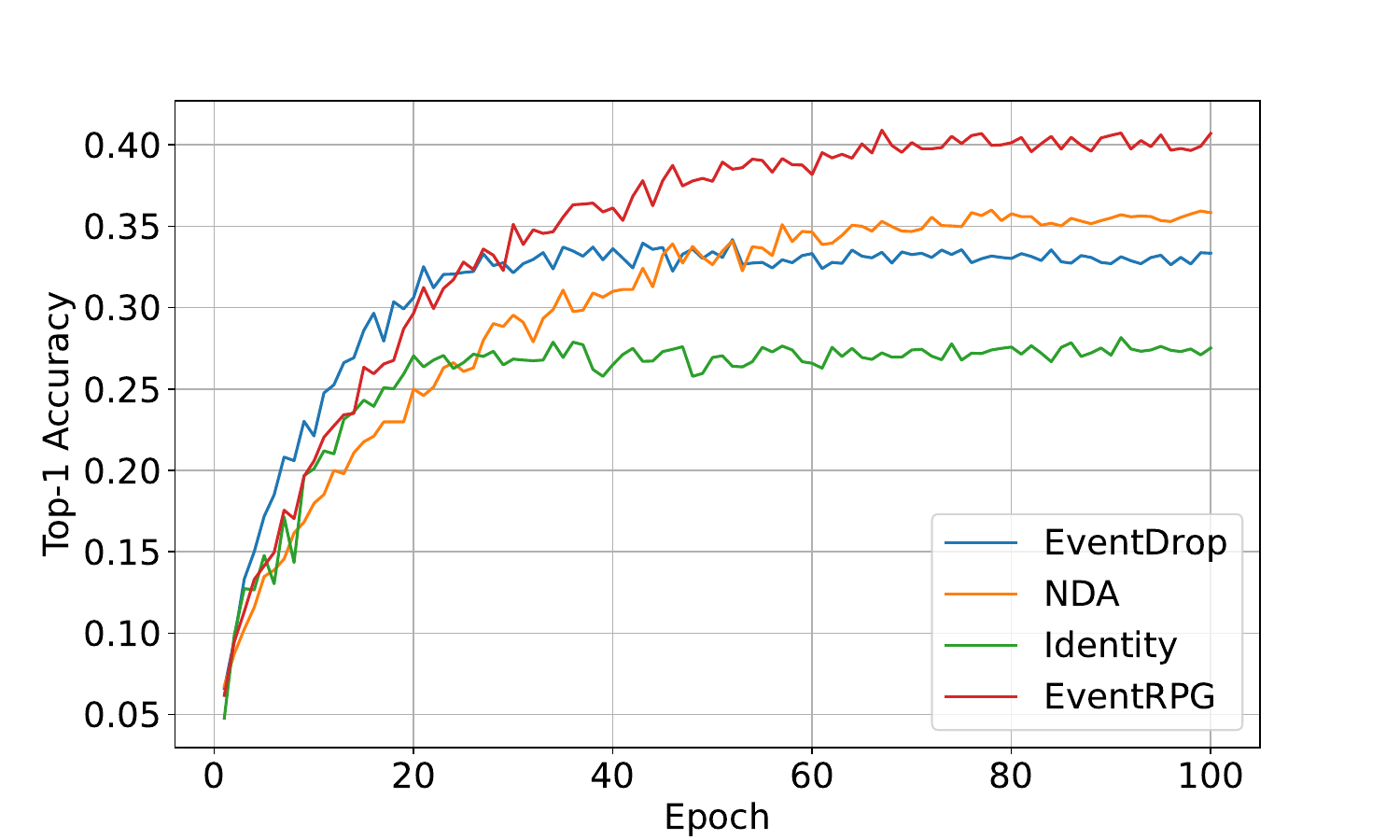}
        \caption{\label{fig:miniNImagenet-acc}}
    \end{subfigure}
    \begin{subfigure}{0.49\textwidth}
      \centering   
      \includegraphics[width=\linewidth]{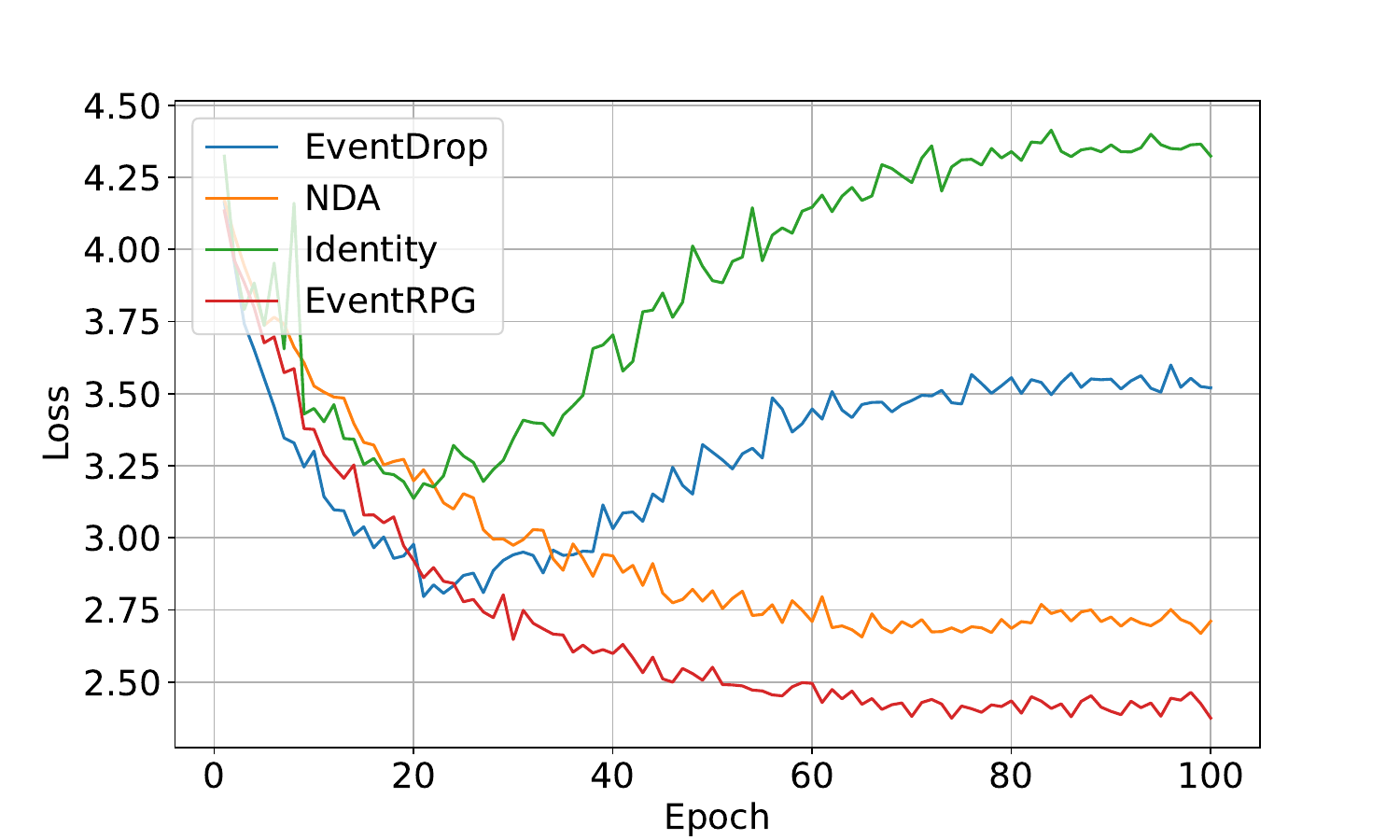}
        \caption{\label{fig:miniNImagenet-loss}}
    \end{subfigure}
\caption{
\label{fig:miniNImagenet}
The loss and Accuracy curves of different augmentations on mini N-ImageNet.}

\end{figure}

As shown in \cref{fig:miniNImagenet-acc}, the performance disparity between \texttt{EventRPG} and NDA is as significant as the one observed between EventDrop and Identity. Additionally, late epochs of EventDrop and Identity exhibit overfitting, a problem that is effectively mitigated in NDA and \texttt{EventRPG}, as illustrated in \cref{fig:miniNImagenet-loss}. These outcomes highlight again the efficiency of our approach in boosting the model's performance and mitigating overfitting.

\subsection{Comparison with other Saliency-based Mix Methods}
We also compare the performance of various saliency-based mix methods on two typical datasets: N-Caltech101 and SL-Animals. In the experients,  there is a $50\%$ chance for each event stream to be augmented by corresponding mix approach with no other augmentations applied. Notably, it's observed in \cref{tab:saliencymix} that both Saliency Mix~\citep{uddin2020saliencymix} and Puzzle Mix~\citep{kim2020puzzle}  occasionally result in lower performance than the identity approach. This underscores RPGMix's robust generalization capability, as it consistently excels across all datasets.

\begin{table}[h]
  \centering
  \resizebox{0.9\textwidth}{!}{
    \begin{tabular}{cc|ccccc} 
      \toprule
      Dataset          & Model                     & Identity & Saliency Mix & Puzzle Mix              & Puzzle Mix (mask only)  & RPGMix  \\
      \midrule
      N-Caltech101     & Spike-VGG11                    & 75.70    & 72.63        & 79.38                   & 78.13                   & \textbf{81.75}   \\
      SL-Animals-4Sets & SEW Resnet18              & 85.42    & 84.00        & 85.34                   & 85.68                   & \textbf{88.67}   \\
      SL-Animals-3Sets & SEW Resnet18              & 89.09    & 89.29        & 88.39                   & 90.18                   & \textbf{90.45}   \\
      \bottomrule
      \end{tabular}
  }
  \caption{Accuracy of Puzzle Mix, Saliency Mix, and RPGMix on object and action recognition tasks.}
  \label{tab:saliencymix}
\end{table}

\begin{figure}[htbp]
  \centering
  \includegraphics[width=\textwidth]{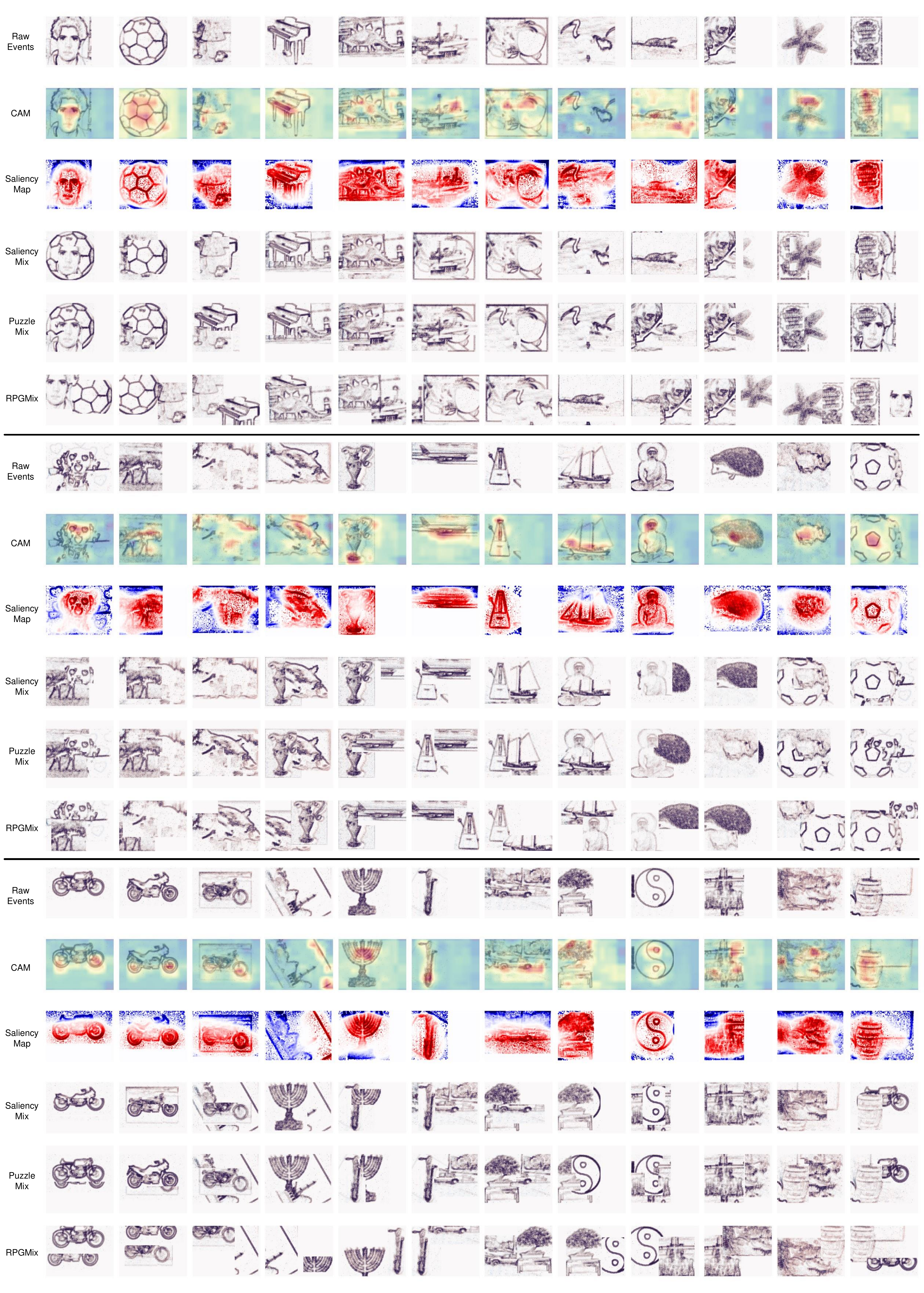}
  \caption{Mixed samples generated by different methods on N-Caltech101.}
  \label{fig:mix_samples}
  \vspace{-0.2cm}
\end{figure}

To clearly showcase the differences among these mix methods, we present visualizations of the augmented samples they produce (see \cref{fig:mix_samples}). In most cases, our RPGMix aims to preserve as many label-related pixels as possible.

\subsection{Additional Quailitative Results}
\begin{figure}[htbp]
  \centering
  \includegraphics[width=\textwidth]{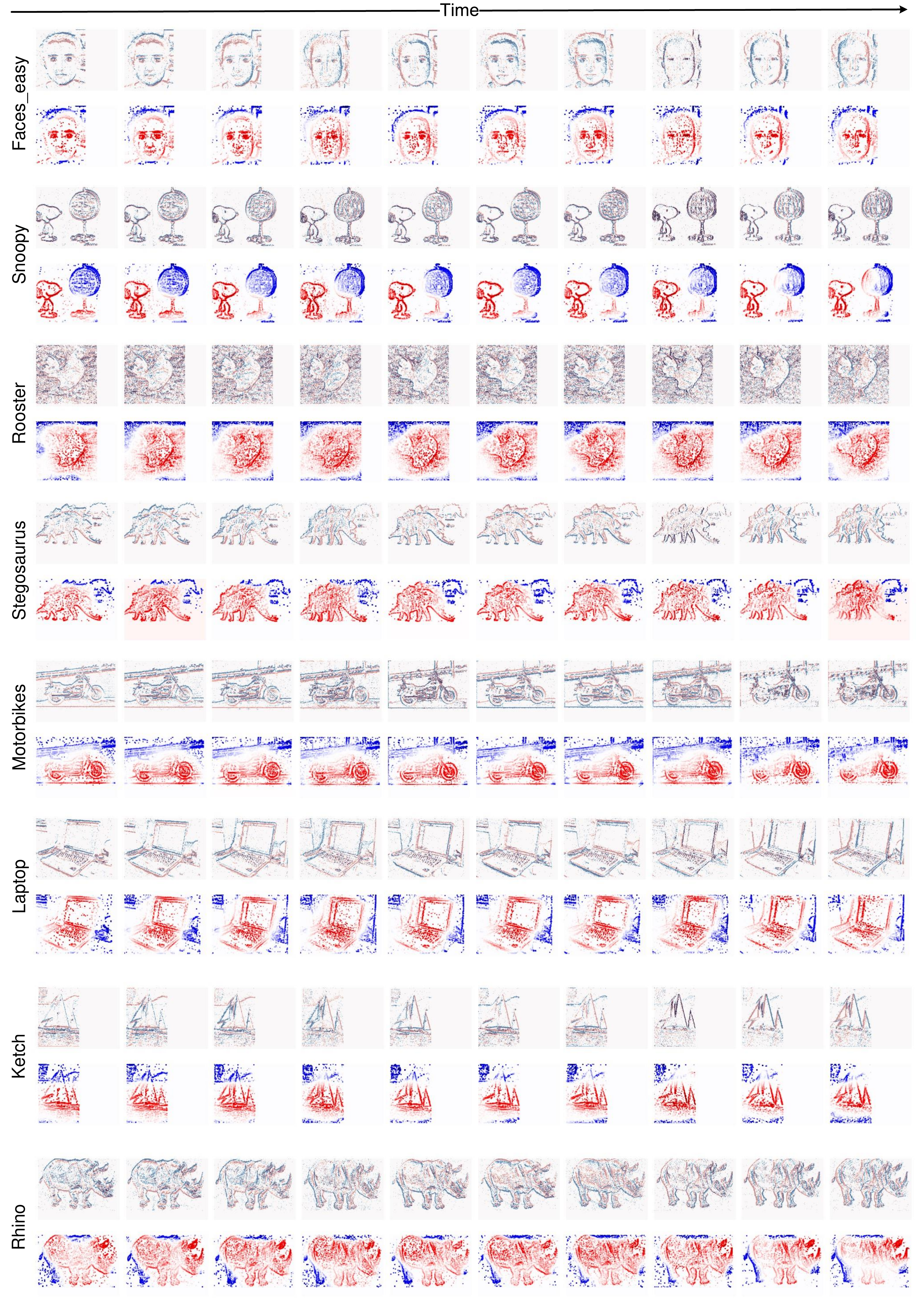}
  \caption{SLTRP-Saliency Maps from Spike-VGG11 on N-Caltech101.}
  \label{fig:ncaltech_sltrp}
  \vspace{-0.2cm}
\end{figure}

\begin{figure}[htbp]
  \centering
  \includegraphics[width=\textwidth]{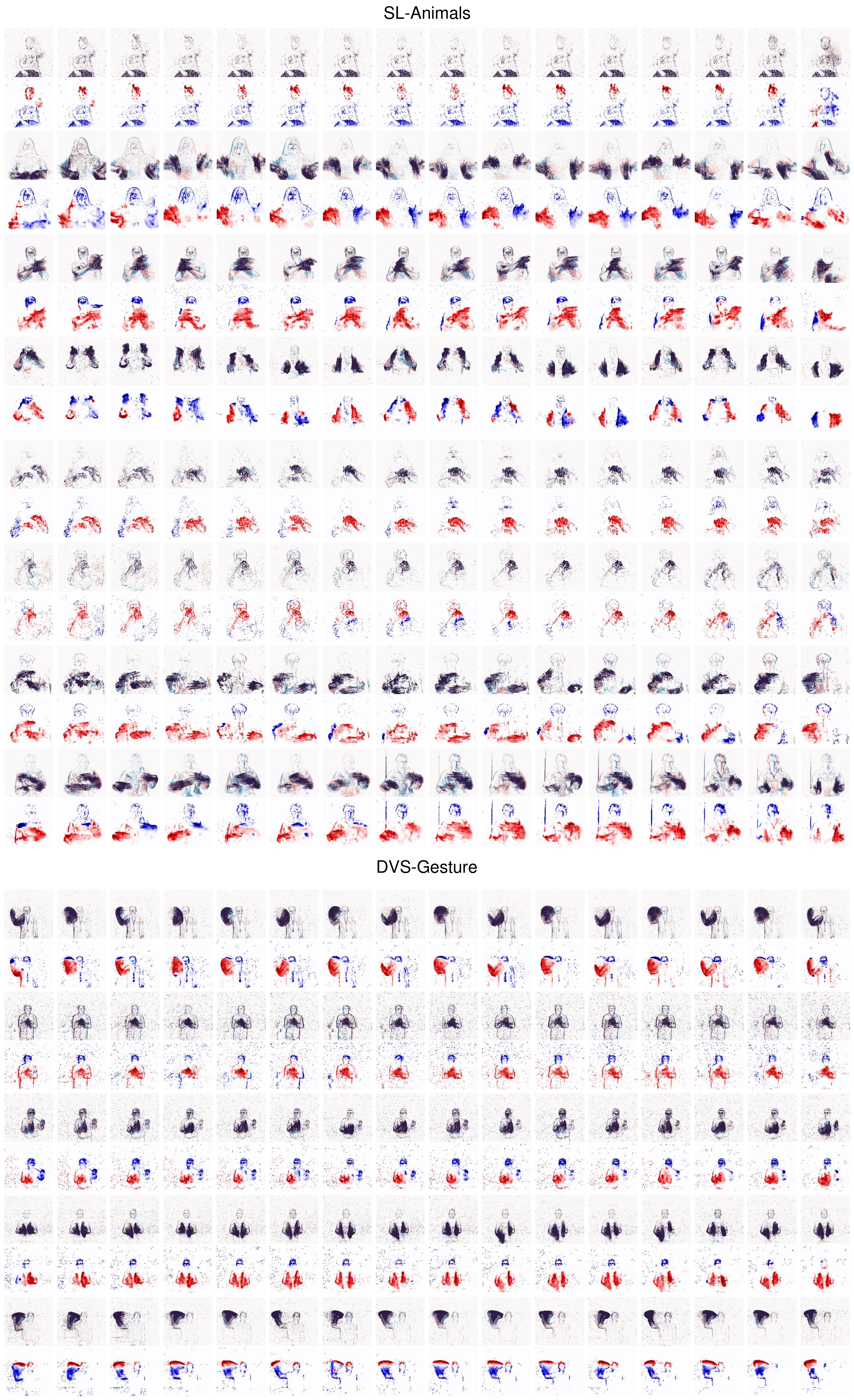}
  \caption{SLTRP-Saliency Maps from SEW Resnet-18 on SL-Animals and DVS-Gesture.}
  \label{fig:action_sltrp}
  \vspace{-0.2cm}
\end{figure}

We present additional qualitative results of our method applied to the N-Caltech101, SL-Animals, and DVS-Gesture datasets. The saliency maps created using \texttt{SLTRP}, as shown in \cref{fig:ncaltech_sltrp}, exhibit strong selectivity. They assign high values to label-related objects and low values to label-unrelated objects. In action recognition datasets, saliency maps primarily focus on label-related moving objects to capture the dynamic aspects of actions, as illustrated in \cref{fig:action_sltrp}. In SL-Animals datasets (second row), the saliency maps focus on the person's hand raised to his/her head, assigning minimal attention to the lower area of the person, despite its high event density. Similarly, in the fourth row, the saliency maps show little interest in the person's right hand (from our perspective), focusing instead on the left hand, even though both hands have similar event densities. These results illustrate that it's the label-related actions, rather than event density, that truly draw the focus of the saliency maps. This observation further confirms the selectivity of SLTRP-generated saliency maps in both action recognition and object recognition datasets.

\section{Related Work}

\subsection{Event-based Data}
Event-based data is generated by asynchronous sensors, usually referred to as event cameras~\citep{mahowald1994silicon,son20174}. Similar to point cloud, event-based data consists of four-dimensional points, denoted as $(x, y, t, p)$, where x and y are the spatial coordinates, t is the timestamp, and p is the polarity. Due to the inherent advantages of event cameras, event-based data is widely used in the field of perception, e.g., optical flow estimation~\citep{gehrig2021raft}, depth estimation~\citep{wang2021stereo}, 3d reconstruction~\citep{baudron2020e3d,rudnev2022eventnerf}, motion segmentation~\citep{stoffregen2019event,zhou2021event}, semantic segmentation~\citep{kim2022beyond}, etc.

There are two mainstream approaches to processing event-based data, frame-based approaches and event-by-event-based approaches. Frame-based approaches~\citep{kogler2009bio,bardow2016simultaneous,lagorce2016hots,gehrig2019end} are similar to image processing, where events are first converted into frames with a fixed shape $(C, H, W)$, and then the frames are fed into an ANN for downstream classification or regression tasks. Event-by-event-based approaches handle events on a one-by-one basis, which is natural for SNNs~\citep{lee2020spike,fang2021deep,deng2022temporal}.

\subsection{Data Augmentation}

Vanilla data augmentations include simple transformations such as scale, rotation, flip, etc. Mixup~\citep{zhang2018mixup} proposes mixing two randomly selected samples and their corresponding labels as new data and label for training, based on which several works that modify the mixing details were proposed to further improve the robustness and performance~\citep{yun2019cutmix,hendrycks2019augmix}. \citet{kim2020puzzle,uddin2020saliencymix} are similar to our proposed RPGMix that leverages the model's saliency information to augment the data, while our motivation and implementation details are quite different from them.

\subsection{Relevance Propagation}
\label{relprop}
Relevance propagation was initially presented in \citep{montavon2017explaining} as a method for creating saliency maps to visualize the impact of each pixel in the input data on the prediction of the model or a particular class, thereby enhancing the interpretability of neural networks. The specificity of saliency maps towards the target class can be improved by employing Contrastive Layer-wise Televance Propagation (CLRP) as described by \citet{gu2018understanding}, which diminishes the relevance scores for non-target classes.

\subsection{Class Activation Map}

As another visualization tool describing the area of most interest to the model, Class Activation Map~(CAM) is widely used to interpret the model's attention and to find objects belonging to a certain class in the input data. The original CAM~\citep{zhou2016learning} could only visualize the activation map of the last global average pooling~(GAP) layer of the model, requiring the model to have a special structure. Grad-CAM~\citep{selvaraju2017grad} overcomes this limitation by replacing the weights of GAP with the sum of gradients, enabling the acquisition of CAM from any CNN model. On this basis, Grad-CAM++~\citep{chattopadhay2018grad}, Score-CAM~\citep{wang2020score}, and Relevance-CAM~\citep{lee2021relevance} were proposed to improve the quality of CAMs.

\section{Datasets and Training Details}
\label{app:dataset}
\subsection{Object Recognition Task}
\paragraph{N-Caltech101} The neuromorphic version~\citep{orchard2015converting} of Caltech101~\citep{fei2004learning}. It is artificially created by moving an asynchronous time-based image sensor~(ATIS) mounted on a pan tilt unit in front of an LCD screen that presents the image data in Caltech101. We use the same dataset as \citet{NDA} that is split into the training set and test set by $9:1$. The resolution of N-Caltech101 is $180\times 240$ which, in our implementation, is padded to $240\times 240$ and rescaled to $128\times 128$.

\paragraph{CIFAR10-DVS} The DVS version~\citep{li2017cifar10} of CIFAR10~\citep{krizhevsky2009learning}. The generation of CIFAR10-DVS is similar to N-Caltech101. Following NDA~\citep{NDA}, we divide it by $9:1$ as the training set and test set, and scale the resolution from $128\times 128$ to $48\times 48$.

\paragraph{N-Cars} A binary classification dataset~\citep{sironi2018hats}. Unlike N-Caltech101 and CIFAR10-DVS transformed from image datasets, N-Cars is obtained from the recording of an ATIS in real driving scenarios. Similar to the preprocessing of N-Caltech101, we first pad the resolution from $100\times 120$ to $120\times 120$ and then scale it to $48\times 48$ on SNN.

\paragraph{Mini N-ImageNet} A subset of N-ImageNet dataset~\citep{kim2021n}. Despite being a smaller segment, it includes over 100,000 samples spanning 100 classes, making it the largest dataset used in our experiments. In line with the original authors' setup, we pad the resolution from $640 \times 480$ to $640 \times 640$ and then scale it down to $224 \times 224$ as the input resolution of the SNN.

\subsection{Action Recognition Task}

For object recognition tasks, shapes and textures are the most important information to recognize an object, and thus we do not care how an object moves in the event streams. While in action recognition tasks, the movements of the objects should be considered. They represent two different recognition strategies and thus are both essential to validate our proposed method.

\paragraph{DVS128 Gesture}
A hand gesture dataset~\citep{dvs_gesture} recorded from a DVS128 camera. It comprises 1464 samples with 11 classes, split into the training set and test set by 8:2. The resolution is set as $128\times 128$ on SNN.

\paragraph{SL-Animals-DVS}
A sign language dataset~\citep{vasudevan2021sl} including 19 Spanish Sign Language signs corresponding to animals. Following the raw paper, we split it by 7.5:2.5 into the training set and test set. Also, we separate the datasets into ``4 sets" and ``3 sets", which excludes the samples disturbed by the indoor lighting conditions reflecting on the patterned clothing of the user. The resolution processing of SL-Animals-DVS is identical to that of DVS 128 Gesture.

\subsection{Training Details}

\begin{table}[]
\small
\centering
\begin{tabular}{ccccccc} 
  \toprule
  Neural Networks                & Neuron Model          & Datasets        & Epoch                & Batch Size          & T                   & Learning Rate                             \\ 
  \midrule
  \multirow{3}{*}{Spike-VGG11}   & \multirow{3}{*}{LIF}  & N-Caltech101    & \multirow{3}{*}{100} & 16                  & \multirow{3}{*}{10} & \multirow{3}{*}{$1\times10^{-3}$}         \\
                                 &                       & CIFAR10-DVS     &                      & 64                  &                     &                                           \\
                                 &                       & N-Cars          &                      & 64                  &                     &                                           \\ 
  \midrule
  \multirow{3}{*}{SEW Resnet-18} & \multirow{3}{*}{PLIF} & SL-Animals      & \multirow{2}{*}{200} & \multirow{2}{*}{20} & \multirow{2}{*}{16} & \multirow{2}{*}{$5\times10^{-4}$}         \\
                                 &                       & DVS-Gesture     &                      &                     &                     &                                           \\
                                 &                       & Mini N-Imagenet & 100                  & 64                  & 10                  & $1\times10^{-3}$  \\
  \bottomrule
  \end{tabular}
\vspace{+0.2cm}
\caption{Hyper-parameters for different models and datasets.}
\label{apptab:1}
\end{table}

In all experiments, we use Adam optimizer with the default setting $(\beta_1, \beta_2)=(0.9,0.999)$ to update the parameters and Cosine Annealing Scheduler for the decrease of learning rate. Other hyper-parameters are shown in \cref{apptab:1}. For fair comparisons, an identical seed is leveraged for all experiments. We utilize the public code of TET~\citep{deng2022temporal} to build Spike-VGG11 and Spikingjelly~\citep{SpikingJelly} to build SEW Resnet-18 with PLIF~\citep{fang2021incorporating}.

\section{Proofs}
In this section, we provide proofs of the propositions.
\subsection{Proof for Conservation Property on $\alpha\beta$-rule}
\label{sec:ab_proof}
\begin{proof}
The sum of relevance scores in layer $l-1$ is
\begin{align*}
\sum_i R^{(l-1)}_i=\sum_i \sum_j R_{i \leftarrow j}^{(l-1, l)}&=\sum_i \sum_j R_{j}^{(l)} \cdot\biggl(\alpha \cdot \frac{z_{i j}^{+}}{\sum_{i}z_{i j}^{+}}+\beta \cdot \frac{z_{i j}^{-}}{\sum_{i}z_{i j}^{-}}\biggr)\\
&=\sum_j R_{j}^{(l)} \sum_i \biggl(\alpha \cdot \frac{z_{i j}^{+}}{\sum_{i}z_{i j}^{+}}+\beta \cdot \frac{z_{i j}^{-}}{\sum_{i}z_{i j}^{-}}\biggr)\\
&=\sum_j R_{j}^{(l)} \biggl(\alpha \cdot \frac{\sum_i z_{i j}^{+}}{\sum_{i}z_{i j}^{+}}+\beta \cdot \frac{\sum_i z_{i j}^{-}}{\sum_{i}z_{i j}^{-}}\biggr)\\
&=\sum_j R_{j}^{(l)} (\alpha+\beta)\\
&=\sum_j R_{j}^{(l)}.
\end{align*}
\end{proof}
\subsection{Proof for \Cref{prop:1}}
\label{sec:prop1proof}
\begin{proof}
We prove \cref{prop:1} by induction.

For case $t=1$, according to \cref{eq:12} we have
\begin{equation*}
R^{(l-1)}[1]=(1-\gamma[1])\left(\sum^T_{i=2}R^{(l)}[i]\prod_{j=2}^{i}\gamma[j]+R^{(l)}[1]\right),
\end{equation*}
where $\gamma[1]=0$, since at the first time step, the membrane voltage is initialized to be $0$. Clearly, \cref{prop:1} holds for case $t=1$.

Next, suppose \cref{prop:1} holds for case $t=k-1$, we have
\begin{align*}
\sum^{k-1}_{t=1} R^{(l-1)}[t] &=\sum^{k-1}_{t=1} R^{(l)}[t] + \sum^T_{i=k}R^{(l)}[i]\prod_{j=k}^{i}\gamma[j]\\
&=\sum^{k-1}_{t=1} R^{(l)}[t] + \gamma[k]\biggl(R^{(l)}[k] + \sum^T_{i=k+1}R^{(l)}[i]\prod_{j=k+1}^{i}\gamma[j] \biggr)\\
&=\sum^{k-1}_{t=1} R^{(l)}[t] + \frac{\gamma[k]}{1-\gamma[k]}R^{(l-1)}[k].
\end{align*}
Then
\begin{align*}
\sum^{k}_{t=1} R^{(l-1)}[t]=\sum^{k-1}_{t=1} R^{(l-1)}[t] + R^{(l-1)}[k]&=\sum^{k-1}_{t=1} R^{(l)}[t] + \frac{\gamma[k]}{1-\gamma[k]}R^{(l-1)}[k] + R^{(l-1)}[k]\\
&=\sum^{k-1}_{t=1} R^{(l)}[t] + \frac{R^{(l-1)}[k]}{1-\gamma[k]}\\
&=\sum^{k-1}_{t=1} R^{(l)}[t] + R^{(l)}[k] + \sum^T_{i=k+1}R^{(l)}[i]\prod_{j=k+1}^{i}\gamma[j]\\
&=\sum^{k}_{t=1} R^{(l)}[t] + \sum^T_{i=k+1}R^{(l)}[i]\prod_{j=k+1}^{i}\gamma[j].
\end{align*}
\end{proof}

\end{document}